\documentclass[lettersize,journal]{IEEEtran}
\usepackage{amsmath,amsfonts}
\usepackage{algorithmic}
\usepackage{algorithm}
\usepackage{array}
\usepackage[caption=false,font=normalsize,labelfont=sf,textfont=sf]{subfig}
\usepackage{textcomp}
\usepackage{stfloats}
\usepackage{url}
\usepackage{verbatim}
\usepackage{graphicx}
\usepackage{booktabs}
\usepackage{threeparttable}
\usepackage{multirow}
\usepackage{subfig}
\usepackage{enumerate}
\usepackage{fontawesome}
\usepackage{hyperref}
\usepackage{tcolorbox}
\usepackage{amssymb}
\usepackage{mathtools}
\usepackage{amsthm}
\usepackage{cite}
\usepackage{makecell}
\usepackage{bm}
\def\BibTeX{{\rm B\kern-.05em{\sc i\kern-.025em b}\kern-.08em
    T\kern-.1667em\lower.7ex\hbox{E}\kern-.125emX}}
\usepackage{balance}

\newtheorem{lemma}{Lemma}
\newtheorem{theorem}{Theorem}
\newtheorem{assumption}{Assumption}
\newtheorem{remark}{Remark}

\makeatletter
\renewcommand\normalsize{%
\@setfontsize\normalsize\@xpt\@xiipt
\abovedisplayskip 4\p@ \@plus3\p@ \@minus3\p@
\abovedisplayshortskip \z@ \@plus3\p@
\belowdisplayshortskip 4\p@ \@plus3\p@ \@minus3\p@
\belowdisplayskip \abovedisplayskip
\let\@listi\@listI}
\makeatother

\usepackage{enumitem}
\setlist[itemize,1]{label=$\bullet$}   
\setlist[itemize,2]{label=$\circ$}     
\setlist[itemize,3]{label=$\star$}     

\allowdisplaybreaks[3]

\begin{document}
\title{FedCanon: Non-Convex Composite Federated Learning with Efficient Proximal Operation on Heterogeneous Data}
\author{Yuan Zhou, Jiachen Zhong, Xinli Shi, \IEEEmembership{Senior Member, IEEE}, \\ Guanghui Wen, \IEEEmembership{Senior Member, IEEE}, and Xinghuo Yu, \IEEEmembership{Fellow, IEEE}
	\thanks{
  This work was supported by the National Natural Science Foundation of China under Grant Nos. 62473098, 62325304, U22B2046, and 62088101, in part by the Australian Research Council under Grant DE250100961. \textit{(Corresponding author: Xinli Shi.)}
		
		Yuan Zhou and Jiachen Zhong are with the School of Cyber Science and Engineering, Southeast University, Nanjing 210096, China (e-mail: zhouxyz@seu.edu.cn; jiachen\_zhong@seu.edu.cn).

        Xinli Shi is with the School of Cyber Science and Engineering and the National Center for Applied Mathematics in Jiangsu, Southeast University, Nanjing,  210096, China, and also with the School of Engineering, RMIT University, Melbourne, VIC 3001, Australia. (e-mail: xinli\_shi@seu.edu.cn).
			

        Guanghui Wen is with the School of Automation, Southeast University, Nanjing 210096, China (e-mail: ghwen@seu.edu.cn).

        Xinghuo Yu is with the School of Engineering, RMIT University, Melbourne, VIC 3001, Australia (e-mail: x.yu@rmit.edu.au).
	}
}

\markboth{FedCanon: Non-{C}onvex Composite Federated Learning with Efficient Proximal Operation on Heterogeneous Data}%
{}

\maketitle

\begin{abstract}
Composite federated learning offers a general framework for solving machine learning problems with additional regularization terms.
However, existing methods often face significant limitations: many require clients to perform computationally expensive proximal operations, and their performance is frequently vulnerable to data heterogeneity.
To overcome these challenges, we propose a novel composite federated learning algorithm called \textbf{FedCanon}, designed to solve the optimization problems comprising a possibly non-convex loss function and a weakly convex, potentially non-smooth regularization term. 
By decoupling proximal mappings from local updates, FedCanon requires only a single proximal evaluation on the server per iteration, thereby reducing the overall proximal computation cost.
Concurrently, it integrates control variables into local updates to mitigate the client drift arising from data heterogeneity. 
The entire architecture avoids the complex subproblems of primal-dual alternatives. 
The theoretical analysis provides the first rigorous convergence guarantees for this proximal-skipping framework in the general non-convex setting. 
It establishes that FedCanon achieves a sublinear convergence rate, and a linear rate under the Polyak-Łojasiewicz condition, without the restrictive bounded heterogeneity assumption. 
Extensive experiments demonstrate that FedCanon outperforms the state-of-the-art methods in terms of both accuracy and computational efficiency, particularly under heterogeneous data distributions.
\end{abstract}

\begin{IEEEkeywords}
Federated Learning, Non-convex Composite Optimization, Convergence Rate, Data Heterogeneity.
\end{IEEEkeywords}

\section{Introduction}
Traditional machine learning often relies on aggregating data for model training, which may raise significant privacy concerns and incur high communication costs, particularly for large-scale learning tasks \cite{mothukuri2021survey}. 
To address these challenges, Federated Learning (FL) has been proposed \cite{huang2024federated}, with Federated Averaging (FedAvg) \cite{mcmahan2017communication} being one of the most commonly utilized frameworks. 
In FedAvg, multiple clients collaboratively train a global model under the coordination of a central server, without sharing their local datasets. 
The framework allows clients to perform multiple rounds of local training before communication.
By effectively leveraging distributed devices, preserving privacy, and reducing communication overhead, FL has achieved significant progress in fields such as computer vision, natural language processing  \cite{bao2022fast,kairouz2021advances,zhu2021federated,valdeira2025communication}.

While most FL research focuses on smooth objectives, additional regularization terms are often essential for obtaining solutions with desirable properties such as sparsity or low rank \cite{xin2021stochastic,yuan2021federated,guo2023decentralized}, leading to the following composite FL formulation:
\begin{equation}\label{equ:problem}
\min_{z \in \mathbb{R}^d} \phi (z) = f(z) + h(z), \ f(z) \triangleq \frac{1}{N} \sum_{i=1}^{N} f_i(z),
\end{equation}
where $z \in \mathbb{R}^d$ is the global model, $N$ denotes the number of clients participating in the training, $f_i$ is the local loss function, and $h$ is a possibly non-smooth regularization term. 
For the composite optimization problem \eqref{equ:problem}, the most commonly used algorithm is Proximal Gradient Descent (PGD), which is a forward-backward splitting method \cite{combettes2011proximal,li2018simple,ryu2022large}:
\begin{equation}\label{equ:PGD}
    z^{t+1} = \textbf{prox}_{\alpha h} \{ z^{t} - \alpha \nabla f(z^{t}) \}, \ t=0,\cdots,T-1,
\end{equation}
where $\textbf{prox}_{\alpha h}\{y\} = \arg\min_{x} \left\{ h(x) + \frac{1}{2\alpha} \|x - y\|^2 \right\}$ represents the proximal operator associated with $h$, and $\alpha>0$ is the step size.
Following the same principle, we adapt FedAvg for composite FL by performing proximal steps locally:
\begin{subequations}\label{equ:FedPGD}
    \begin{align}
&\hat x^{t,0}_i=z^t, \ \hat x^{t,k+1}_i=\textbf{prox}_{\beta h}\{\hat x^{t,k}_i - \beta g_i(\hat x^{t,k}_i) \},  \label{equ:FedPGD1}\\
    &\bar\Delta^t=\frac{1}{N}\sum_{i=1}^N\Delta_i^t=\frac{1}{N}\sum_{i=1}^N\left(z^t-\hat x^{t,K}_i\right), \label{equ:FedPGD2}\\
    &z^{t+1} = \textbf{prox}_{\alpha h} \{ z^{t} -  \alpha{\bar\Delta^t} \},\label{equ:FedPGD3}
\end{align}
\end{subequations}
where $\hat x^{t,k}_i$ represents the local model, and $g_i(\hat x^{t,k}_i)$ is the unbiased and variance-bounded stochastic gradient.
The above \eqref{equ:FedPGD1} corresponds to the local updates of client $i$ for $k=0,\cdots,K-1$.
Meanwhile, the server also maintains the global gradient $z^t$, which is broadcast at the start of the local updates. 
Afterwards, the server uses the aggregated result from \eqref{equ:FedPGD2} to replace the gradient $\nabla f(z^{t})$ in \eqref{equ:PGD} and update the global model in \eqref{equ:FedPGD3}.
{Essentially, \eqref{equ:FedPGD} can be viewed as FedMiD \cite{yuan2021federated} under the standard Euclidean geometry, where the general mirror descent step simplifies to a proximal update.}
However, directly using \eqref{equ:FedPGD} to solve  \eqref{equ:problem} may lead to the following three potential issues.

\textbf{P1: The Curse of Primal Averaging}.
Information aggregated by the server typically consists of post-proximal local states. 
Since proximal mappings are typically nonlinear, \eqref{equ:FedPGD2} cannot be telescoped as in FedAvg to extract the gradient $\frac{1}{NK}\sum_{i=1}^N\sum_{k=0}^{K-1}g_i(\hat x^{t,k}_i)$ \cite{zhang2024composite}. 
Moreover, averaging locally sparse models (e.g., induced by $\ell_1$-norm) often results in a dense global model, a phenomenon known as the ``curse of primal averaging" \cite{yuan2021federated}.
To address this, \cite{yuan2021federated} proposes an FL algorithm based on Dual Averaging (FedDA), where the dual states are aggregated. 
Meanwhile, the algorithm in \cite{zhang2024composite} (denoted as ZA1) decouples the proximal operator from communication and requires clients to directly transmit the pre-proximal local models.
These allow the server to recover the gradient information effectively. 
However, with the notable exception of several primal-dual methods (e.g., FedDR \cite{tran2021feddr}, FedADMM \cite{wang2022fedadmm}), existing composite FL algorithms  \cite{bao2022fast,zhang2024composite,yuan2021federated} conduct convergence analyses that predominantly rely on (strong) convexity assumptions.
This limits their applicability to the non-convex problems typical in deep learning.

\textbf{P2: Client-Side Computational Bottleneck}.
Directly applying \eqref{equ:FedPGD} imposes a heavy computational load because each client must execute $K$ proximal mappings per round, which is a common requirement in existing composite FL methods \cite{bao2022fast,yuan2021federated,zhang2024composite}.
Although many commonly used regularization terms have fixed and known proximal mappings, this does not necessarily imply computational efficiency.
As shown in TABLE \ref{table:3}, for deep models like ResNet \cite{he2016deep}, the cost of a single proximal operation can be comparable to, or even exceed, that of a full forward-backward pass.
It is therefore evident that the time spent handling the regularization term becomes significant and cannot be overlooked.
Fortunately, \cite{mishchenko2022proxskip}  proposes ProxSkip, which allows for optionally skipping the proximal mapping step after a gradient descent update, providing critical inspiration for the subsequent research, though it is designed for centralized, strongly convex settings.
Although extended to decentralized optimization in \cite{guo2023revisiting}, the ProxSkip framework (and its federated adaptation SCAFFNEW \cite{mishchenko2022proxskip,guo2023revisiting}) is not directly applicable to the general composite FL problem \eqref{equ:problem}. 
Specifically, SCAFFNEW simplifies the problem by treating the non-smooth term as the server's aggregation-averaging step, effectively reducing it to a smooth optimization task rather than solving the general composite objective.


\begin{table}[tbp]
\setlength{\tabcolsep}{2pt}
\centering
{
\caption{
Time Cost (milliseconds) and Relative Overhead of Proximal Mappings on Different Neural Networks.
}\label{table:3}
\begin{threeparttable}
\begin{tabular}{c|c|ccc}
\toprule
{Models} &Fwd \& Bwd & $\ell_1$-Norm & MCP & SCAD \\
\midrule
ResNet-18 & 86.6 & 119.8 (1.383$\times$) & 222.4 (2.568$\times$)  & 387.4 (4.473$\times$)  \\
ResNet-34 & 159.1 &226.8 (1.426$\times$)  &371.9 (2.338$\times$)  &751.9  (4.727$\times$) \\
ResNet-50 &302.5 &248.3 (0.821$\times$) & 401.8 (1.328$\times$) &836.2 (2.764$\times$)  \\
ResNet-101 &561.3 &430.0 (0.766$\times$) & 718.7 (1.280$\times$) & 1589.1 (2.831$\times$) \\
\bottomrule
\end{tabular}
\begin{tablenotes}[flushleft]
\item  All elements in the table are tested on the same device, with the MNIST dataset. 
The model size increases from top to bottom.
``MCP" refers to Minimax Concave Penalty, and ``SCAD" refers to Smoothly Clipped Absolute Deviation.
\end{tablenotes}
\end{threeparttable}
}
\end{table}

\textbf{P3: Client Drift under Data Heterogeneity}.
Due to the distributed storage nature of FL, data heterogeneity is inevitable, often manifesting as label, feature, or quantity skew \cite{kairouz2021advances,pei2024review,zhu2021federated,liu2022fedbcd}. 
This heterogeneity causes local updates to diverge significantly, a phenomenon known as ``client drift", which negatively impacts convergence \cite{mendieta2022local,wang2021novel,jhunjhunwala2022fedvarp,karimireddy2020scaffold}.
Most existing analyses rely on the restrictive assumption that data heterogeneity is bounded (i.e., bounded gradient dissimilarity) \cite{karimireddy2020scaffold,liconvergence,cheng2023momentum,xiangefficient,yu2019parallel}.
This assumption appears in various forms, as detailed in Section \ref{sec:ProblemFormulation}.
While strategies like incorporating global gradients \cite{karimireddy2020scaffold,mishchenko2022proxskip,zhang2024composite} or adding penalty terms \cite{durmus2021federated,zhang2021fedpd} can mitigate drift, they are predominantly designed for smooth optimization. 
Extending such drift-correction mechanisms to the non-convex composite setting without relying on bounded heterogeneity assumptions remains a significant challenge.


Primal-dual algorithms like FedDR \cite{tran2021feddr}, FedADMM \cite{wang2022fedadmm}, offer a structured approach to these challenges. 
By reformulating the problem with consensus constraints, they decouple the objective, allowing the smooth loss and the non-smooth regularization term to be handled separately by clients and the server, respectively. 
By excluding the proximal mapping of $h$ from local updates, this architecture inherently circumvents P1 and P2, and the use of dual variables and proximal penalty terms provides robustness against client drift (P3).
However, primal-dual methods face a practical hurdle: they require clients to solve local Lagrangian subproblems to high accuracy each round \cite{li2019communication}. 
In stochastic settings, this is often infeasible and inevitably introduces inexactness errors. 
More critically, these errors can accumulate across the network, potentially degrading the final accuracy of the global model. 
This motivates the search for a simple and robust primal-only algorithm that avoids such subproblem-solving complexities.

\begin{table*}[htbp]
\renewcommand\arraystretch{1.15}
\centering
\caption{
Comparison of Some Related Works on Composite Optimization or Data Heterogeneity.
}\label{table:compare}
\begin{threeparttable}
\begin{tabular}{c|cccccc}
\toprule
\multirow{2}{*}{\textbf{Methods}} &  \multicolumn{2}{c}{\textbf{Optimization Can Be}}   &\textbf{Proximal}  & \textbf{Robust to Data} & \textbf{Simple}   & \textbf{Linear Rate}    \\
 & \textbf{Composite?} & \textbf{Non-convex?}   &  \textbf{Efficiency?}  & \textbf{Heterogeneity?} &  \textbf{Subproblems?} & \textbf{Condition}   \\
\midrule
 FedAvg-M \cite{cheng2023momentum}   & \faTimes & \faCheck &\faTimes & \faCheck  &\faCheck &\faTimes \\
 SCAFFOLD \cite{karimireddy2020scaffold}   & \faTimes & \faCheck &\faTimes & \faCheck  &\faCheck & Strong Convexity  \\
{ProxSkip \cite{mishchenko2022proxskip}}  & \faCheck & \faTimes & \faCheck &\faCheck &\faCheck & Strong Convexity  \\
 {SCAFFNEW \cite{mishchenko2022proxskip}}  & \faTimes & \faTimes & \faTimes &\faCheck &\faCheck &  Strong Convexity  \\
    Algorithm in \cite{guo2023revisiting} &\faTimes &\faCheck &\faTimes &\faCheck &\faCheck &Strong Convexity \\
  FedMiD/FedDA \cite{yuan2021federated}   &\faCheck &\faTimes &\faTimes &\faTimes &\faCheck &\faTimes  \\ 
   Fast-FedDA \cite{bao2022fast}   &\faCheck &\faTimes &\faTimes &\faTimes &\faCheck &\faTimes  \\ 
  Algorithm in \cite{zhang2024composite} &\faCheck &\faTimes &\faTimes &\faCheck &\faCheck & Strong Convexity \\
       FedPD \cite{zhang2021fedpd} &\faTimes &\faCheck &\faTimes &\faCheck & \faTimes &\faTimes \\
     FedDyn \cite{durmus2021federated} &\faTimes &\faCheck &\faTimes &\faCheck &\faTimes & Strong Convexity \\
   FedDR \cite{tran2021feddr} &\faCheck &\faCheck &\faCheck &\faCheck &\faTimes &\faTimes \\
   FedADMM \cite{wang2022fedadmm} &\faCheck &\faCheck &\faCheck &\faCheck &\faTimes &\faTimes  \\
\midrule
 FedCanon in This Paper  &  \faCheck & \faCheck & \faCheck & \faCheck &\faCheck & PL Condition  \\
\bottomrule
\end{tabular}
\end{threeparttable}
\end{table*}

\textbf{Contributions.}
{In response to these problems, we propose FedCanon, a composite \underline{Fed}erated learning algorithm with effi\underline{C}ient proxim\underline{a}l operatio\underline{n}s and the ability to address data heter\underline{o}ge\underline{n}eity.
Our main contributions are threefold:}
{
\begin{itemize}
  \item \textbf{An Efficient Primal-Only Framework}. 
    By decoupling proximal operators from local updates, FedCanon  circumvents the curse of primal averaging (P1) and reduces the proximal computation from $\mathcal{O}(NK)$ to a single operation per round---marking the first application of proximal-skipping to non-convex composite FL (P2). 
    Furthermore, it integrates control variables to correct for client drift (P3) without relying on bounded heterogeneity assumptions.
    This primal-only design avoids the complex subproblem solving inherent in primal-dual methods, offering a simple yet effective update rule.
    To our knowledge, FedCanon is the first to simultaneously achieve computational efficiency, robustness to heterogeneity, and applicability to non-convex composite problems within a simple primal-only structure.
  \item \textbf{Rigorous Convergence Guarantees under General Assumptions.} 
We provide the first  convergence analysis for an FL algorithm combining efficient proximal operations and control variables in the general non-convex composite setting.
This analysis is non-trivial because the radical design introduces a \textbf{structural drift}: client updates, lacking information about the non-smooth term $h$, systematically mismatch the global objective.
This is a challenge that conventional methods are specifically designed to avoid.
Unlike primal-dual methods that couple the local and global models via dual variables, our analysis directly confronts  this discrepancy to maintain a simple solver.
Without restrictive assumptions, we prove that the proximal gradient norm converges at a sublinear rate of $\mathcal{O}(1/T)$ to a steady-state error. 
Under the Polyak-Łojasiewicz (PL) condition, FedCanon converges linearly to a neighborhood of the optimal solution.
  \item \textbf{Comprehensive Numerical Experiments.} 
  We apply FedCanon to real neural networks and training datasets, and compare its performance with several state-of-the-art FL methods. 
  Experimental results show that, FedCanon effectively handles proximal operations under heterogeneous data distributions, achieving shorter training time and higher accuracy, thereby validating its effectiveness.
\end{itemize}}

To highlight the main features of this paper, we present a comparison of related works in TABLE \ref{table:compare}.

\textbf{Notation.}
In the paper, we symbolize the $d$-dimensional vector space as $\mathbb{R}^{d}$ and use $\mathbf{0}_d$ to represent a $d$-dimensional zero vector. 
The $\ell_1$-norm and $\ell_2$-norm are denoted by $\Vert \cdot \Vert_1$ and $\Vert \cdot \Vert$, respectively. 
We use $[N]$ to represent the set $\{1, \cdots , N\}$. 
The differential operator is denoted by $\nabla$, the subdifferential operator by $\partial$, and expectation by $\mathbb{E}$.
Additionally, $\mathbb{N}_+$ denotes the set of all positive integers.

\section{Proposed Algorithm}
In this section, we will first present the problem formulation along with some necessary assumptions, followed by a detailed explanation of the construction process and discussions.

\subsection{Problem Formulation}\label{sec:ProblemFormulation}
We consider the composite optimization problem in FL of the following form:
\begin{equation}\label{equ:problem_consensus}
\min_{\{x_i\}_{i=1}^N, z} \  \frac{1}{N} \sum_{i\in[N]} f_i(x_i) + h(z), \  s.t. \  x_i = z, \ \forall i \in [N], 
\end{equation}
where $x_i\in\mathbb{R}^d, i\in [N]$ are local copies of the global variable $z\in\mathbb{R}^d$. 
Fundamentally, \eqref{equ:problem_consensus} is equivalent to \eqref{equ:problem} due to the consensus constraint.
The local loss function $f_i(x_i):=\mathbb{E}_{\xi_i\sim\mathcal{D}_i}[F_i(x_i;\xi_i)]$ is a smooth, potentially non-convex function, where $\mathcal{\xi}_{i}$ is an independent sample drawn from the local data distribution $\mathcal{D}_i$ of client $i$. 
No extra assumptions are made on the similarity of local data distributions, which highlights the presence of data heterogeneity.
The function $f$ can capture a wide range of functions commonly used in neural networks, such as sigmoid function, while $h$ represents a possibly non-smooth regularization term (e.g., $\ell_1$ norm, MCP, SCAD penalty) or an indicator function over a closed and compact set.

We now provide more detailed assumptions below.
\begin{assumption}\label{assum:1}
The objective function $\phi$ is lower bounded, i.e., $\phi^* = \inf_z \phi (z) > -\infty$.
\end{assumption}
\begin{assumption}\label{assum:2}
The local function $f_i : \mathbb{R}^d \mapsto \mathbb{R}$ is $L$-smooth, i.e., for any $x, y\in\mathbb{R}^d$, there exists $L > 0$ such that
\begin{subequations}
    \begin{align}
    &f_i(x) - f_i(y) - \langle \nabla f_i(y), x-y \rangle \le \frac{L}{2} \Vert x - y \Vert^2, \label{equ:lipschitz} \\
    &\Vert\nabla  f_i(x)-\nabla  f_i(y)\Vert\leq L\Vert x-y\Vert. \label{equ:lipschitz2}
\end{align}
\end{subequations}
\end{assumption}
Assumptions \ref{assum:1} and \ref{assum:2} are standard conditions commonly adopted in the convergence analysis for non-convex optimization   \cite{tran2021feddr,zhang2021fedpd}. 
The smoothness condition in Assumption \ref{assum:2} is weaker than the sample-wise smoothness condition required in \cite{cheng2023momentum,mancino2023proximal,liu2022fedbcd,xin2021stochastic}.

\begin{assumption}\label{assum:4}
The function $h:\mathbb{R}^d\rightarrow\mathbb{R}\cup\{+\infty\}$ is proper, closed and $\rho$-weakly convex (i.e., $h(z)+\frac{\rho}{2}{\Vert z\Vert}^2$ is convex with $\rho \ge 0$), but not necessarily to be smooth.
It typically represents a regularization term scaled by a factor $\kappa>0$.
Moreover, for any $h'(z)\in\partial h(z)$, there exists a constant $B_h =\mathcal{O}(\kappa)$ such that $\Vert h'(z)\Vert^2 \le B_h$.
\end{assumption}

Weak convexity generalizes standard convexity, reducing to it when $\rho = 0$.
Although weakly convex functions are non-convex, they retain several valuable properties of convex functions (see Lemma \ref{lem:weakly_convex} for details). 
For instance, the proximal operator $\textbf{prox}_{\alpha h}$ remains applicable to weakly convex functions, provided that $0 < \alpha < 1/\rho$.
{Common examples include the MCP and SCAD penalty. 
Unlike the $\ell_1$ norm which can introduce significant estimation bias by over-penalizing large coefficients, these non-convex penalties are designed to be nearly unbiased for large parameters, often leading to improved model accuracy and sparsity patterns \cite{bohm2021variable,yan2023compressed}.}
A broader class of regularization terms with bounded subgradients is discussed in \cite{xu2023comparative}.
In practice, these regularization terms are typically scaled by a factor, denoted as $\kappa$, to appropriately balance their influence. 
For example, when employing the $\ell_1$ norm, we have $h(z) = \kappa \Vert z\Vert_1$ with $B_h = \kappa$. 
In the training of large-scale models, $\kappa$ is often chosen to be small in order to mitigate instability during optimization.
Compared to existing works that typically assume convex regularization terms, e.g., \cite{mancino2023proximal,yan2023compressed,yuan2021federated,zhang2024composite}, Assumption \ref{assum:4} is more general and easily satisfied, thereby broadening the applicability of algorithm.


\begin{assumption}\label{assum:5}
If $\{\xi_{i,b}\}_{b=1}^B$ are independent samples from $\mathcal{D}_i$ with $B \in \mathbb{N}_+$, the local stochastic gradient estimator $g_i(x_i;\xi_i) := \frac{1}{B} \sum_{b=1}^B \nabla F_{i}(x_i; \xi_{i,b})$ satisfies \cite{xin2021stochastic}:
\begin{subequations}\label{equ:minibatch}
\begin{align}
&\mathbb{E}[g_i(x_i;\xi_i)]=\nabla f_i(x_i), \label{equ:minibatch1} \\
&\mathbb{E}{\Vert g_i(x_i;\xi_i) - \nabla f_i(x_i)\Vert}^2\leqslant {\sigma^2/B}. \label{equ:minibatch2}
\end{align} 
\end{subequations}
{For brevity, we write $g_i(x_i) :=g_i(x_i;\xi_i)$ and suppress $\xi_i$ in the notation henceforth.}
\end{assumption}
Due to limited computational resources of devices, clients in FL typically use unbiased and variance-bounded stochastic gradients instead of the true gradients for large-scale training tasks.
To further reduce the variance, we employ the above stochastic gradient estimator based on batch samplings.

In addition to these, many FL-related works also introduce bounded heterogeneity assumptions. 
These assumptions typically require a certain relationship between global and local gradients\footnote{Although the first condition only requires the maximum local gradient to be bounded, it implies ${\|\nabla f(x) \|}^2={\|\frac{1}{N}\sum_{j\in[N]}\nabla f_j(x) \|}^2\!\leq\! \frac{1}{N}\sum_{j\in[N]}{\|\nabla f_j(x) \|}^2\!\leq\! \zeta^2$. This implies that the bounded gradient assumptions can be viewed as  special cases of bounded heterogeneity assumptions.
}, which can take the following forms \cite{xiangefficient,bian2024accelerating,yuan2021federated}:
\begin{align*}
    &\max_{\forall j\in[N]} {\|\nabla f_j(x)\|}^2 \leq \zeta^2;\ \frac{1}{N}\sum_{j\in[N]}{\|\nabla f(x) - \nabla f_j(x)\|}^2 \leq \zeta^2; \\
    &\qquad\qquad\frac{1}{N}\sum_{j\in[N]}{\|\nabla f_j(x)\|}^2 \leq \tau^2{\|\nabla f(x)\|}^2 +\zeta^2.
\end{align*}
In contrast, the proposed FedCanon in this paper does not require any such assumptions.

\subsection{The FedCanon Algorithm}

\begin{algorithm}[tbp]
\caption{FedCanon}
\label{alg:FedCanon}
\begin{algorithmic}
\STATE {\bfseries Input:} Step size parameters $\alpha , \beta > 0$, number of iterations $T \ge 1$, number of local updates $K \ge 1$, initial global variable $z^{0}\in \mathbb{R}^d$,  initial control variables $\{c_i^0\}_{i\in[N]} \in \mathbb{R}^d$ with $\sum_{i\in[N]} c_i^0 = \mathbf{0}_d$. 
\FOR {$t = 0,1,\cdots,T-1$} 
    \STATE \begin{tcolorbox}[colback=green!5!white, colframe=green!55!black, boxrule=0.5pt,width={200pt},top={-3pt},bottom={-1pt},left={3pt},title={\textbf{On Client} $i\in [N]$ \textbf{in Parallel Do}}]
        \STATE Set $\hat{x}_i^{t, 0} = z^{t}$.
        \FOR{$k = 0,1, \cdots ,K-1$}
            \STATE Update $\hat{x}_i^{t, k+1} = \hat{x}_i^{t, k} - \beta\left[g_i(\hat{x}_i^{t, k})+c_i^t\right]$.
        \ENDFOR
        \STATE \textbf{Send} $\Delta_i^{t+1} = \frac{1}{\beta K} (z^{t} - \hat{x}_i^{t, K})$ to server.
         \STATE \textbf{Receive} $(\bar{\Delta}^{t+1}, z^{t+1})$ from server. 
         \STATE {Update} $c_i^{t+1} = c_i^t+\bar{\Delta}^{t+1}-\Delta_i^{t+1}$.
         \end{tcolorbox}
    \STATE \begin{tcolorbox}[colback=blue!5!white, colframe=blue!60!white, boxrule=0.5pt,width={200pt},top={-3pt},bottom={-1pt},left={3pt},title={\textbf{On Server Do}}]
    \STATE \textbf{Receive} ${\Delta}_i^{t+1}$ from all clients. 
        \STATE {Update} $\bar{\Delta}^{t+1} = \frac{1}{N} \sum_{i\in[N]} \Delta_i^{t+1}$.
        \STATE {Update} $z^{t+1} = \textbf{prox}_{\alpha h} \{ z^{t} - \alpha \bar{\Delta}^{t+1} \}$.
        \STATE \textbf{Broadcast} $(\bar{\Delta}^{t+1}, z^{t+1})$ to all clients.\end{tcolorbox}
\ENDFOR
\end{algorithmic}
\end{algorithm}

The complete procedure of FedCanon is outlined in Algorithm \ref{alg:FedCanon}. 
This algorithm involves $T$ rounds of communications between clients and the server, where each client performing $K$ local updates during each round. 
To distinguish between the two levels of iteration, superscripts $t$ and $k$ are used to denote global and local iterations, respectively.
At initialization, the step sizes $\alpha$ and $\beta$, as well as the number of global iterations $T$ and local updates $K$, are specified. 
The server initializes the global variable $z^0$ and broadcasts it to all clients. 
Each client then initializes its local control variable $c_i^0$ such that $\sum_{i\in[N]} c_i^0 = \mathbf{0}_d$ (the simplest approach is to set $c_i^0 = \mathbf{0}_d$ for all clients).  
During the $t$-th iteration, client $i$ initializes its local model $\hat{x}_i^{t, 0}$ using the global model $z^t$ received from the previous round, and then performs $K$ local updates. 
In each local update, client $i$ computes a batch-sampled stochastic gradient $g_i(\hat{x}_i^{t, k})$ and adjusts it using its local control variable $c_i^t$ to alleviate client drift. 
After completing $K$ local updates, client $i$ calculates the difference between its initial and updated local model, sending $\Delta_i^{t+1}$ to the server.
The server aggregates $\Delta_i^{t+1}$ from all clients and computes their average $\bar{\Delta}^{t+1}$. 
Using this aggregated gradient estimate, the server performs a proximal gradient descent step on $z^t$, yielding the updated global variable $z^{t+1}$.
Finally, the server broadcasts $(\bar{\Delta}^{t+1}$, $z^{t+1})$ to all clients, enabling them to update their local models and control variables in preparation for the next round.

The main steps of FedCanon are summarized as follows:
\begin{subequations}\label{equ:iteration}
\begin{align}
\hat{x}_i^{t, 0} &= z^{t}, \label{equ:iteration0}\\
\hat{x}_i^{t, k+1} &= \hat{x}_i^{t, k} - \beta\left[g_i(\hat{x}_i^{t, k})+c_i^t\right], \ k=0,\cdots, K-1, \label{equ:iteration1}\\
\Delta_i^{t+1} &= \frac{1}{\beta K} \left(z^{t} - \hat{x}_i^{t, K}\right) = v_i^{t} + c_i^t, \label{equ:iteration2} \\
\bar{\Delta}^{t+1} &=\frac{1}{N}\sum_{j\in[N]}\Delta_j^{t+1}=\bar{v}^{t},  \label{equ:iteration3}\\
z^{t+1} &= \textbf{prox}_{\alpha h} \{ z^{t} - \alpha \bar{\Delta}^{t+1} \},  \label{equ:iteration4}\\
c_i^{t+1} &= c_i^t + \bar{\Delta}^{t+1} - \Delta_i^{t+1} = \bar{v}^{t} - v_i^{t} \label{equ:iteration5} \\
&=c_i^t + \frac{1}{\beta NK} \sum_{j\in[N]}\left(z^{t} - \hat{x}_j^{t, K}\right)-\frac{1}{\beta K} \left(z^{t} - \hat{x}_i^{t, K}\right) \notag\\
&=c_i^t -\frac{1}{\beta K}\big(\frac{1}{N}\sum_{j\in[N]}\hat{x}_j^{t, K}-\hat{x}_i^{t, K}\big),    \notag
\end{align} 
\end{subequations}
where $v_i^{t} = \frac{1}{K} \sum_{k=0}^{K-1} g_i(\hat{x}_i^{t, k})$ and $\bar{v}^{t} = \frac{1}{N} \sum_{j\in[N]} v_j^{t}$. 
The last equation in \eqref{equ:iteration2} is derived by iterating local updates \eqref{equ:iteration1}, while \eqref{equ:iteration3} and \eqref{equ:iteration5} hold due to $$\frac{1}{N}\sum_{j\in[N]}c_j^{t+1}=\frac{1}{N}\sum_{j\in[N]}c_j^{t}=\cdots=\frac{1}{N}\sum_{j\in[N]}c_j^{0}=\mathbf{0}_d.$$

The structure of this algorithm is simple, and follows the general framework of classical algorithms such as FedAvg and SCAFFOLD.
However, it is specifically designed to handle more general non-convex composite optimization problems. 
A key feature of FedCanon is that it delegates the computation of the proximal mapping to the server side, performing this operation only once per communication round, as shown in \eqref{equ:iteration4}.
This undoubtedly reduces the computational cost significantly, compared with FedMiD and FedDA in \cite{yuan2021federated}, ZA1 in \cite{zhang2024composite}, particularly in large-scale scenarios with frequent proximal computations on clients.

\subsection{A Variant for Communication-Constrained Scenarios}
Compared with SCAFFNEW, Algorithm \ref{alg:FedCanon} requires the server to broadcast more variables to all clients in each communication round. 
To further reduce the number of transmitted variables, modifications can be made such that the proximal mapping step \eqref{equ:iteration4} is performed locally by all clients in parallel, i.e.,
$$\hat{x}_i^{t+1, 0} = \textbf{prox}_{\alpha h}\{\hat{x}_i^{t, 0} - \alpha \bar{\Delta}^{t+1}\}, \ \forall i\in[N].$$ 
In this variant, the server is only responsible for aggregating $\Delta_i$ and broadcasting $\bar\Delta$ to all clients.
If all clients are initialized with the same local model, consistency can be maintained at the beginning of each round, without the need to receive the global model, i.e., $\hat{x}_1^{t, 0} = \cdots = \hat{x}_N^{t, 0}=z^t$. 
We refer to this modified version as \textbf{FedCanon II}, and its detailed steps are provided in Algorithm \ref{alg:FedCanon2}. 
The communication cost of FedCanon II is comparable to that of SCAFFNEW, requiring only two variables to be exchanged between the server and each client per iteration.

However, this reduction in communication cost comes at the expense of increased computational burdens: the number of proximal mappings per iteration increases to $N$, as each client independently performs this step.
{It demonstrates a critical trade-off between communication and computation, and the optimal choice depends heavily on the specific deployment scenario. We analyze two representative cases below:
\begin{itemize}
    \item \textbf{Cross-Silo FL} (e.g., federated training across organizations): 
    In this setting, clients are typically data centers with substantial computational power \cite{kairouz2021advances}, making the local proximal operation very fast.
    The main bottleneck is often the communication burden  over a wide-area network. 
    By eliminating the need to broadcast the updated global model, FedCanon II directly reduces the total server-to-client communication volume by one-third. 
    For large-scale models, this substantial reduction makes it a highly efficient choice, particularly when network bandwidth is a bottleneck.
\item \textbf{Cross-Device FL} (e.g., training on mobile phones): 
In this scenario, clients possess limited computational power \cite{mcmahan2017communication}. 
Forcing them to perform even a single proximal operation can be a significant burden (as shown in Table \ref{table:3}). 
Therefore, offloading this expensive task to the powerful central server, as done in FedCanon (Algorithm \ref{alg:FedCanon}), remains a more practical and resource-efficient approach.
\end{itemize}}
{We conclude that while Algorithm \ref{alg:FedCanon} is generally better suited for resource-constrained cross-device applications, Algorithm \ref{alg:FedCanon2} is a powerful alternative for communication-constrained environments, especially when training large models.}
TABLE \ref{table:compare2} compares the two versions of FedCanon with several existing composite FL methods, reporting the total number of proximal operations and the floats exchanged during communication per iteration.
Notably, when the number of local updates $K$ is large, FedCanon II still preserves the advantage of efficient proximal operations, as it avoids multiple proximal computations on clients during local updates.


\begin{algorithm}[tbp]
\caption{FedCanon II}\label{alg:FedCanon2}
\begin{algorithmic}
\STATE {\bfseries Input:} Step size parameters $\alpha , \beta > 0$, number of iterations $T \ge 1$, number of local updates $K \ge 1$, initial variables $\hat{x}_1^{0, 0}=\cdots=\hat{x}_N^{0, 0}\in \mathbb{R}^d$,  initial control variables $\{c_i^0\}_{i\in[N]} \in \mathbb{R}^d$ with $\sum_{i\in[N]} c_i^0 = \mathbf{0}_d$.
\FOR {$t = 0,1,\cdots,T-1$} 
\STATE \begin{tcolorbox}[colback=green!5!white, colframe=green!55!black, boxrule=0.5pt,width={210pt},top={-3pt},bottom={-1pt},left={3pt},title={\textbf{On Client} $i\in [N]$ \textbf{in Parallel Do}}]
         \FOR{$k = 0,1, \cdots ,K-1$}
            \STATE Update $\hat{x}_i^{t, k+1} = \hat{x}_i^{t, k} - \beta\left[g_i^t(\hat{x}_i^{t, k})+c_i^t\right]$.
        \ENDFOR
        \STATE \textbf{Send} $\Delta_i^{t+1} = \frac{1}{\beta K} (\hat{x}_i^{t, 0} - \hat{x}_i^{t, K})$ to server.
         \STATE \textbf{Receive} $\bar{\Delta}^{t+1}$ from server. 
          \STATE Update $\hat{x}_i^{t+1, 0} =  \textbf{prox}_{\alpha h}\{\hat{x}_i^{t, 0}-\alpha \bar{\Delta}^{t+1}\}$.
         \STATE {Update} $c_i^{t+1} = c_i^t+\bar{\Delta}^{t+1}-\Delta_i^{t+1}$.\end{tcolorbox}
          \STATE \begin{tcolorbox}[colback=blue!5!white, colframe=blue!60!white, boxrule=0.5pt,width={210pt},top={-3pt},bottom={-1pt},left={3pt},title={\textbf{On Server Do}}]
    \STATE \textbf{Receive} ${\Delta}_i^{t+1}$ from all clients. 
    \STATE \textbf{Broadcast} $\bar{\Delta}^{t+1}= \frac{1}{N} \sum_{i\in[N]} \Delta_j^{t+1}$ to clients.\end{tcolorbox}
    \ENDFOR
\end{algorithmic}
\end{algorithm}
\begin{table}[thbp]
\setlength{\tabcolsep}{14pt}
\centering
\caption{
Comparison of Composite FL methods: Total Proximal Mappings and Per-Client Communication Per Iteration.
}\label{table:compare2}
\begin{threeparttable}
\begin{tabular}{c|cc}
\toprule
\multirow{2}{*}{\textbf{Methods}} &  \textbf{Total Proximal} & \textbf{Floats}  \\
&  \textbf{Computations}   & \textbf{Exchange} \\
\midrule
FedCanon & $1$  & $3d$ \\
FedCanon II & $N$ & $2d$  \\
FedMiD / FedDA & $|\mathcal{S}|K+1$ & $2d$  \\
Fast-FedDA &$NK_t+1$ & $4d$\\
 ZA1 & $N(K+1)+1$ & $2d$ \\
\bottomrule
\end{tabular}
\begin{tablenotes}[flushleft]
\item  In this table, we set the model dimension for all algorithms as $d$, where $|\mathcal{S}|$ represents the number of sampled clients, and $K_t$ denotes the number of local updates performed in the $t$-th iteration.
\end{tablenotes}
\end{threeparttable}
\end{table}

\section{Convergence Analysis}
In this section, we will present the convergence rates of FedCanon when solving general non-convex composite optimization problems and those satisfying the PL condition.

We first introduce several commonly used notations for the subsequent analysis.
From the update of $z^{t+1}$ \eqref{equ:iteration4}, we can express it in an alternative form:
\begin{equation}\label{equ:prox_iteration}
z^{t + 1}= z^{t} - \alpha \underbrace{\left[ \frac{1}{\alpha} \left(z^{t} - \textbf{prox}_{\alpha h} \{ z^{t} - \alpha \bar\Delta^{t+1} \} \right) \right]}_{G^{\alpha}(z^{t}, \bar\Delta^{t+1})} .
\end{equation} 
It is worth noting that if $\bar\Delta^{t+1}$ in $G^{\alpha}(z^{t}, \bar\Delta^{t+1}\!)$ is replaced by $\nabla \!f(z^t)$, the resulting expression denotes the proximal gradient:
$$G^{\alpha}(z^{t})\triangleq\frac{1}{\alpha}\left(z^{t} - \textbf{prox}_{\alpha 
h}\{z^{t} - \alpha \nabla f(z^{t})\}\right).$$ 
It is evident that $G^{\alpha}(z^*) = \mathbf{0}_d$ is equivalent to
\begin{equation*}
\mathbf{0}_d \in \nabla f(z^*) + \partial h(z^*),
\end{equation*}
and we refer $z^*$ as the stationary point.
Moreover, $G^{\alpha}(z)$ would reduce to $\nabla f(z)$ when $h=0$.
Therefore, the proximal gradient can be used as a criterion for measuring convergence in some composite optimization algorithms \cite{li2018simple,wang2019spiderboost}.
Next, for $t = 0, \ldots, T-1$, we define the following term:
$$\mathcal{E}^{t+1}=\frac{1}{N}\sum_{i\in[N]} \mathbb{E}{\Vert \nabla f_i(\hat{x}_i^{t+1, 0}) - v_i^{t} - \nabla f({z}^{t+1}) + \bar{v}^{t} \Vert^2}.$$

Then, the subsequent two lemmas all hold for the sequence $\big\{( \{\hat x_i^{t,k}\}_{k=0}^{K-1},\Delta_i^{t+1},c_i^{t+1})_{i=1}^N,(\bar{\Delta}^{t+1},z^{t+1})\big\}_{t=0}^{T-1}$ generated by \eqref{equ:iteration} (equivalent to Algorithms \ref{alg:FedCanon} and \ref{alg:FedCanon2}).
Their detailed proofs are provided in Appendices \ref{appendix_lemma1} and \ref{appendix_lemma2}, respectively.
\begin{lemma} \label{lem:lemma1}
Suppose that Assumptions \ref{assum:1}-\ref{assum:5} hold, if $0<\alpha<\frac{1}{\rho}$ and $\beta^2\le \frac{1}{24K(K-1)L^2}$,  it has
\begin{align} \label{equ:phi_iteration3}
&\mathbb{E}\left[\phi({z}^{t+1}) - \phi({z}^{t})\right] \\
{\le}& - \frac{\alpha\!-\!2(\rho\!+\!L)\alpha^2}{4} \mathbb{E}\Vert G^{\alpha}(z^{t}, \bar\Delta^{t+1}) \Vert^2 -\frac{\alpha}{8}\mathbb{E}\Vert G^{\alpha}(z^{t}) \Vert^2 \notag \\
&-\left[\frac{\alpha}{16}-12(2 + \delta)\alpha\beta^2K^2L^2\right]\mathbb{E}\Vert \nabla f(z^{t})\Vert^2+\frac{\alpha B_h}{8} \notag\\ 
&+{12(2 + \delta)\alpha\beta^2K^2L^2}   \mathcal{E}^t  + \frac{\alpha(2 + \delta)(1+3\beta^2 K^3L^2)\sigma^2}{BK},\notag
\end{align}
where $\delta=1/(1-\alpha\rho)^2$.
\end{lemma}
\begin{lemma} \label{lem:lemma2}
Suppose that Assumptions \ref{assum:1}-\ref{assum:5} hold, then the following inequality holds:
\begin{align}\label{equ:gradient_track}
\mathcal{E}^{t+1}{\le}& {48\beta^2K^2L^2}(\mathcal{E}^t+\mathbb{E}{\Vert \nabla f({z}^{t}) \Vert^2}) \\
&\!+\!2\alpha^2L^2 \mathbb{E}\Vert G^{\alpha}(z^{t}, \bar\Delta^{t+1}) \Vert^2  \!+\! \frac{4\sigma^2(1\!+\!3\beta^2K^3L^2)}{BK}. \notag
\end{align}
\end{lemma}

\begin{remark}
In Lemma \ref{lem:lemma1}, we apply the inequality
$$-\Vert G^{\alpha}(z^{t}) \Vert^2 \le B_h- \frac{1}{2}\Vert \nabla f(z^{t})\Vert^2$$
to introduce a negative term involving $\mathbb{E}\Vert \nabla f(z^{t})\Vert^2$, which naturally motivates the inclusion of $B_h$ in Assumption \ref{assum:4}.
Fortunately, as previously discussed, for many widely used regularization terms, an upper bound on their subgradients can be readily determined. 
Without Assumption \ref{assum:4}, controlling $\mathbb{E}\Vert \nabla f(z^{t})\Vert^2$ requires more restrictive conditions, such as assuming bounded gradients of the loss function, which is an assumption adopted in certain analyses of FedAvg \cite{liconvergence,yu2019parallel}.
\end{remark}

\subsection{Convergence Analysis under General Non-Convex Settings}
By leveraging Lemmas \ref{lem:lemma1} and \ref{lem:lemma2}, it is easy to derive the following Theorem \ref{thm:convergence} for FedCanon applied to general non-convex composite optimization problems.
Specifically, adding $\alpha(\mathcal{E}^{t+1}-\mathcal{E}^t)$ to the left-hand side of \eqref{equ:phi_iteration3} and substituting \eqref{equ:gradient_track} into it, we adjust the parameters $\alpha$ and $\beta$ such that the coefficients of $\mathcal{E}^t$, $\mathbb{E}\Vert \nabla f(z^{t})\Vert^2$ and $\mathbb{E}\Vert G^{\alpha}(z^{t}, \bar\Delta^{t+1}) \Vert^2$ on the right-hand side become negative to eliminate these terms.
By summing and telescoping the resulting inequalities over $t = 0$ to $T-1$, simplifying the expression, and taking the average yield \eqref{equ:convergence}.
The details are presented in Appendix \ref{appendix_theorem1}.
\begin{theorem}\label{thm:convergence}
Suppose that Assumptions \ref{assum:1}-\ref{assum:5} hold, if $0<\alpha(\rho+L)+4\alpha^2L^2 \le \frac{1}{2}$ and $ 192(6+\delta)\beta^2K^2L^2\le 1$, for the sequence generated by Algorithms \ref{alg:FedCanon} and \ref{alg:FedCanon2}, we have
\begin{align}\label{equ:convergence}
\frac{1}{T}\sum_{t=0}^{T-1} \mathbb{E}\Vert G^{\alpha}(z^{t}) \Vert^2 \le& \frac{8[\phi ({z}^{0}) - \phi^{*}+\alpha\mathcal{E}^0]}{\alpha T}  \\
&+\frac{50\sigma^2}{BK}+\frac{\sigma^2}{8B}+B_h. \notag
\end{align}
\end{theorem}
Theorem \ref{thm:convergence} establishes that when the bound $B_h$ can be scaled by $\kappa$, the proximal gradient decreases at a sublinear rate of $\mathcal{O}(1/T)$ until it reaches a steady-state error $\mathcal{O}(\frac{\sigma^2}{BK})+\mathcal{O}(\frac{\sigma^2}{B})+\mathcal{O}(\kappa)$, which is primarily determined by the variance of the stochastic gradients and the boundedness of the subgradient of the regularization term.
This residual error can be further reduced by increasing the mini-batch size $B$ or decreasing the regularization scaling factor $\kappa$ within $h$.
If ${1}/{B}$ and $\kappa$ are both set to $\mathcal{O}(1/T)$, then Theorem \ref{thm:convergence} implies the existence of $0 \le j < T$ such that 
$\mathbb{E}\Vert G^{\alpha}(z^{j}) \Vert^2 \le \epsilon$,
which means FedCanon can obtain an expected $\epsilon$-stationary point with iteration complexity $\mathcal{O}(1/\epsilon)$.

\begin{remark}
To ensure model accuracy, the scaling factor $\kappa$ should not be set too large, especially when training deep neural networks with a large number of parameters.
According to our empirical observations in the next section, when using ResNet architectures, it is advisable to set $\kappa$ no greater than $10^{-5}$.
\end{remark}

\begin{remark}
\textcolor{brown}{
While mega-batches (or additional variance reduction mechanisms) appear theoretically necessary for FedCanon to attain arbitrary precision, this requirement should be viewed in the context of our general assumptions. 
Many existing composite FL methods that allow residual errors to be controlled by step sizes often rely on stronger assumptions, such as (strong) convexity or bounded gradients/heterogeneity \cite{yuan2021federated,bao2022fast}. 
To the best of our knowledge, achieving arbitrary precision under the general non-convex and unbounded heterogeneity setting within a simple primal-only framework remains an open challenge.
}
\end{remark}

\subsection{Convergence Analysis under the PL Condition}
The PL condition is an additional assumption considered in many works for convergence analysis, as it enables linear rate for some non-convex algorithms \cite{karimi2016linear,wang2019spiderboost,yi2022communication}. 
This condition means that every stationary point is a global minimizer, but it is weaker than the $\mu$-strong convexity because the global minimum is not unique. 
Since this work focuses on composite FL, we provide its more general form.
\begin{assumption}[\cite{li2018simple}]\label{ass:pl_condition}
For any $z \in \mathbb{R}^d$, the objective function $\phi$ satisfies the more general PL condition with $\mu > 0$, i.e.,
\begin{equation}\label{equ:pl_defn}
\Vert G^{\alpha}(z) \Vert^2 \ge 2\mu [\phi(z)-\phi^{*}].
\end{equation}
\end{assumption}
In Theorem \ref{thm:pl_convergence}, we present the convergence rate of FedCanon under the additional Assumption \ref{ass:pl_condition}. 
The process of deriving \eqref{equ:pl_convergence} is similar to \eqref{equ:convergence}. 
Specifically, adding $\alpha\mathcal{E}^{t+1}$ to the left-hand side of \eqref{equ:phi_iteration3} and substituting \eqref{equ:gradient_track}, we leverage the inequality from the PL condition to replace the $\mathbb{E}\Vert G^{\alpha}(z^{t}) \Vert^2$ term. 
By adjusting the parameters $\alpha$ and $\beta$ to eliminate the $\mathbb{E}\Vert \nabla f(z^{t})\Vert^2$ and $\mathbb{E}\Vert G^{\alpha}(z^{t}, \bar\Delta^{t+1}) \Vert^2$ terms and ensuring that the coefficient of $\mathcal{E}^t$ does not exceed $\alpha(1-\frac{\alpha\mu}{4})$, we iteratively substitute the simplified inequality backward until $t=0$, which leads to \eqref{equ:pl_convergence}.
The detailed proof is provided in Appendix \ref{appendix_theorem2}.
\begin{theorem}\label{thm:pl_convergence}
Suppose that Assumptions \ref{assum:1}-\ref{ass:pl_condition} hold, if $0<\alpha(\rho+L)+4\alpha^2L^2 \le \min\left\{\frac{1}{2}, \frac{4\mu(\rho+L)+64L^2}{\mu^2}\right\}$ and ${12(6+\delta)\beta^2K^2L^2}\le \min\left\{\frac{1}{16},1-\frac{\alpha\mu}{4}\right\}$, for the sequence generated by Algorithms \ref{alg:FedCanon} and \ref{alg:FedCanon2}, it holds that
\begin{align}\label{equ:pl_convergence}
&\mathbb{E}\left[\phi ({z}^{T})-\phi^*\right] +\alpha\mathcal{E}^{T}  \\
\le&\left(1-\frac{\alpha\mu}{4}\right)^{T} \left[\phi ({z}^{0})-\phi^* +\alpha\mathcal{E}^{0}\right]+ \frac{25\sigma^2}{\mu BK}+ \frac{\sigma^2}{16\mu B}+\frac{B_h}{2\mu}.  \notag 
\end{align}
\end{theorem}
Theorem \ref{thm:pl_convergence} implies that when the objective function satisfies the PL condition, FedCanon can achieve a linear convergence rate with a residual $\frac{1}{2\mu}$-times that of \eqref{equ:convergence}.

\begin{remark}
    {A typical design in FedCanon is the normalized update $\Delta_i^{t+1}=\frac{1}{\beta K} (z^{t} - \hat{x}_i^{t, K})$ transmitted by clients, which is equivalent to the time-average of their local update directions. 
This design decouples the local step size $\beta$ from the global update process. 
Consequently, $\beta$ serves primarily as a local parameter to ensure client-side stability, while $\alpha$ drives global convergence. 
This decoupling implies that the restrictive theoretical condition on $\beta$ does not inherently limit the global convergence rate, as a more aggressive $\alpha$ can be chosen to accelerate training. 
It also provides the rationale for the effective heuristic $\alpha=\beta K$, which links the global update to the cumulative local computation.
}
\end{remark}

\section{Numerical Experiments}\label{section_experiments}
In this section, we evaluate the convergence and effectiveness of FedCanon under different hyperparameter settings and compare it with other state-of-the-art composite FL methods.
All these experiments are executed on a dedicated computing platform equipped with an Intel\textregistered\ Xeon\textregistered\ Gold 6242R CPU, 512GB of RAM and three NVIDIA Tesla P100 GPUs. 
The operating system is Ubuntu 20.04, and the software environment dependencies include Python 3.8 and PyTorch 1.13.

\subsection{Experimental Setup} 

\subsubsection{Problems} 
We set the local loss function as $$f_i(x_i)=\frac{1}{m_i}\sum_{j=1}^{m_i}\mathcal{L}(\mathcal{M}(x_i, a_{i,j}), b_{i,j}),$$ where $\mathcal{M}$ represents the model output when given model parameters $x_i$ and sample $a_{i,j}$, $\mathcal{L}$ represents the cross-entropy loss between $\mathcal{M}(x_i, a_{i,j})$ and the label $b_{i,j}$. 
Meanwhile, the regularization term $h$ is set to $\ell_1$-norm, MCP or SCAD penalty. 
Unless specified, we default to using train loss and test accuracy to illustrate the performance of algorithms.

\subsubsection{Datasets} 
{We employ A9A, MNIST, FMNIST, FEMNIST, CIFAR-10, CINIC-10 \cite{darlow2018cinic}, and TinyImageNet \cite{le2015tiny} datasets for experiments.} 
Model training and accuracy testing are conducted using the pre-split training and validation sets from each dataset. 
These datasets are widely used in the fields of distributed optimization and machine learning, with details summarized in TABLE \ref{table:1}.

\subsubsection{Partitions} 
To investigate the impact of data heterogeneity, we adopt the strategy in \cite{ye2023heterogeneous} to simulate non-i.i.d. (independent and identically distributed) data across clients, which is activated by sampling from a Dirichlet distribution to induce label skew. 
Specifically, we use $Dir(\eta)$ to quantify the degree of data heterogeneity with concentration parameter $\eta > 0$. 
Smaller $\eta$ indicates stronger heterogeneity, while i.i.d. corresponds to an infinite parameter.
{For the FEMNIST dataset, which has a natural user-based structure, we apply the Dirichlet distribution over the writer IDs. 
This creates a more practical user distribution skew, where each client is assigned a varying number of writers' data, naturally inducing both feature and label heterogeneity due to different handwriting styles and character frequencies.
For others, we use the Dirichlet distribution to create label distribution skew, where each client is allocated a skewed distribution of class labels.}

\subsubsection{Models} 
We assess the algorithmic performance of four models: Linear, MLP, CNN, and ResNet-18 \cite{he2016deep}. 
The Linear model represents a simple linear architecture. 
The Multilayer Perceptron (MLP) model consists of three linear layers, each followed by a ReLU activation function. 
To achieve better performance on image data, we employ a convolutional neural network (CNN), which includes two convolutional layers, each followed by a ReLU activation function and a max-pooling layer. 
ResNet-18, a more complex yet powerful model, is used for further experiments. 

\subsubsection{Algorithms} 
We select FedAvg \cite{mcmahan2017communication}, SCAFFOLD \cite{karimireddy2020scaffold}, SCAFFNEW \cite{mishchenko2022proxskip}, FedMiD \cite{yuan2021federated}, FedDA \cite{yuan2021federated}, ZA1 \cite{zhang2024composite}, FedDR \cite{tran2021feddr} and FedADMM \cite{wang2022fedadmm} as the baseline algorithms for comparison with FedCanon.

\begin{table}[htbp]
\centering
\caption{An Overview of the Employed Datasets.}\label{table:1}
\begin{tabular}{c|ccc}
\toprule
\textbf{Datasets} & \textbf{Train / Eval Samples} & \textbf{Classes} & \textbf{Input Size} \\
\midrule
A9A & 32,561 / 16,281 & 2 & 123 \\
MNIST / FMNIST & 60,000 / 10,000 & 10 & 1$\times$28$\times$28 \\
{FEMNIST} & {712,825 / 92,438} & {62} & {1$\times$28$\times$28} \\
CIFAR-10 & 50,000 / 10,000 & 10 & 3$\times$32$\times$32 \\
{CINIC-10} & {90,000 / 90,000} & {10} & {3$\times$32$\times$32} \\
{TinyImageNet}  & {100,000 / 10,000} & {200} & {3$\times$64$\times$64} \\
\bottomrule
\end{tabular}
\end{table}

\subsection{Experimental Results}

\begin{figure*}[htbp]
    \centering
    \subfloat{\includegraphics[width=0.25\linewidth]{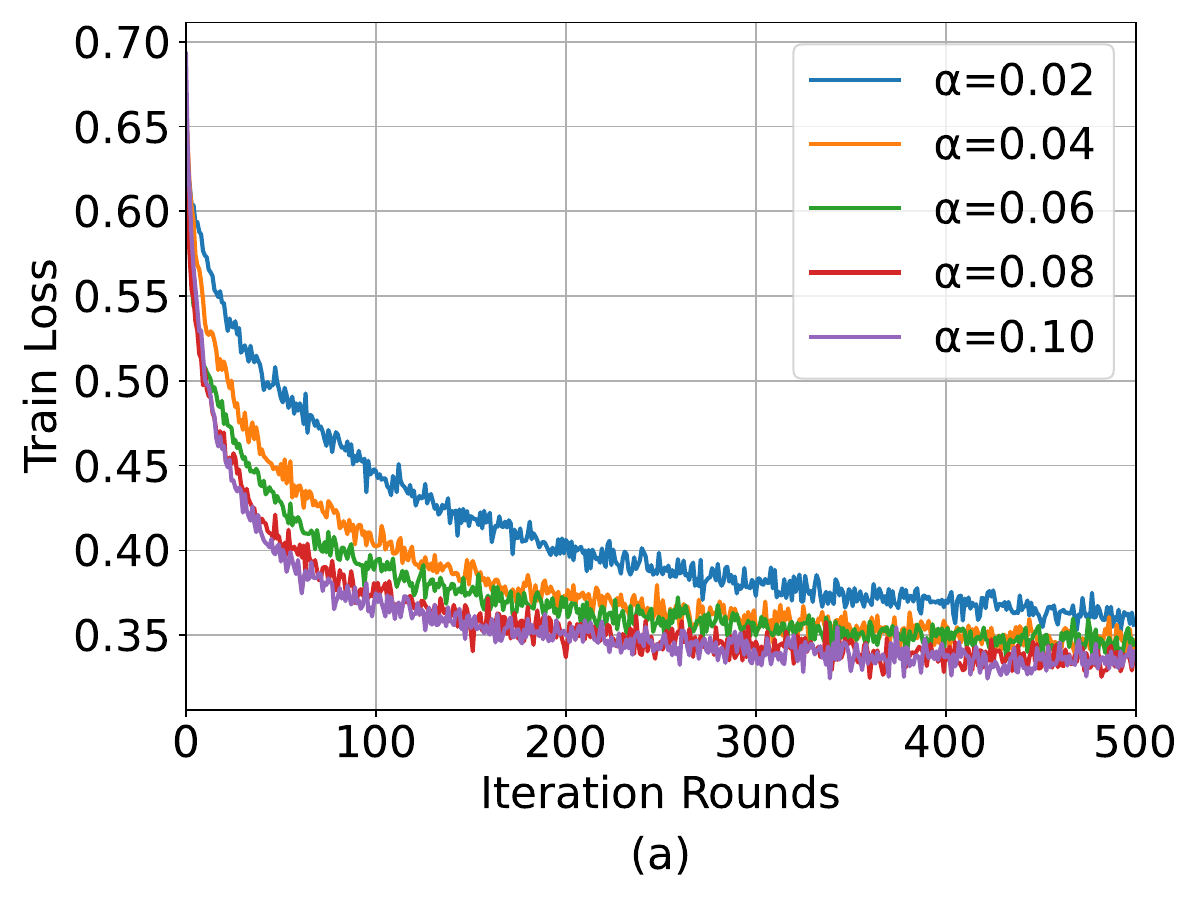}}
    \subfloat{\includegraphics[width=0.25\linewidth]{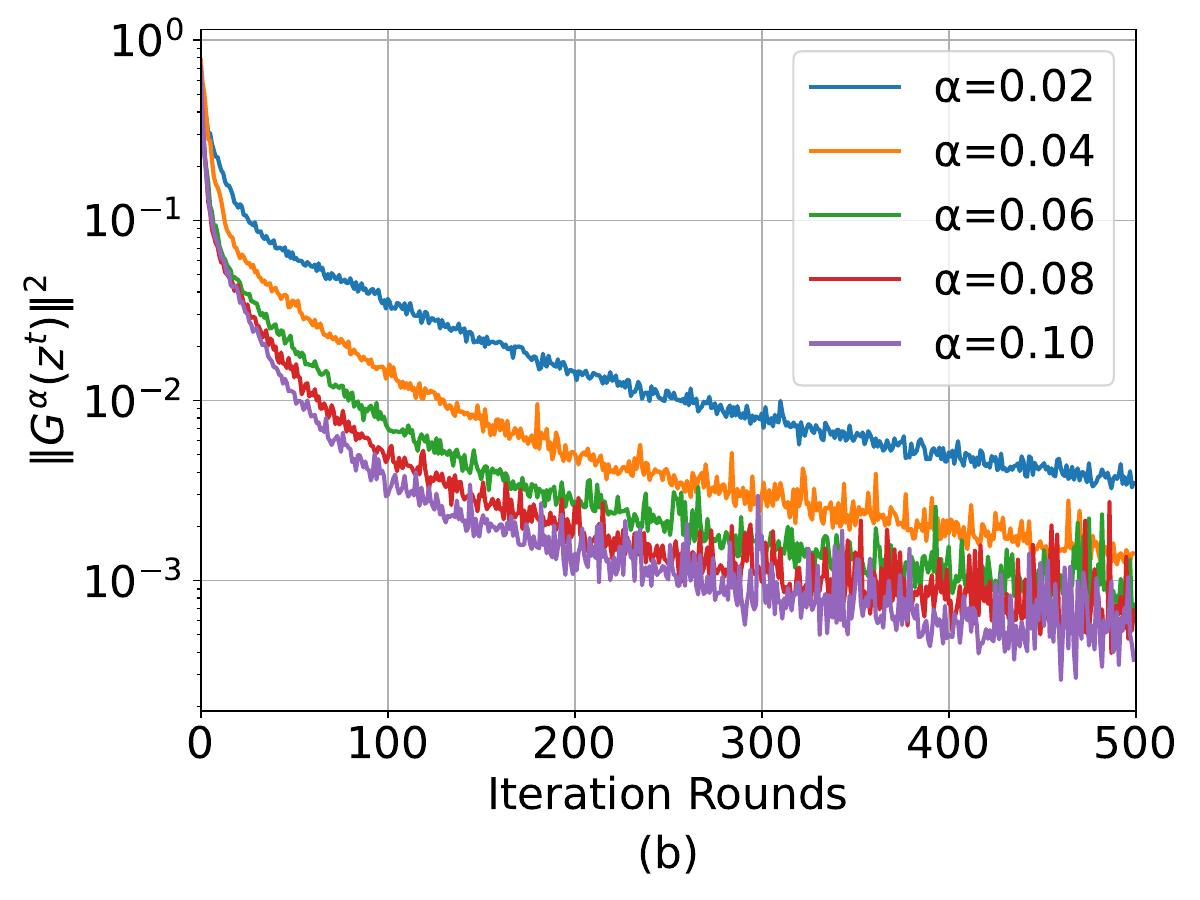}}
    \subfloat{\includegraphics[width=0.25\linewidth]{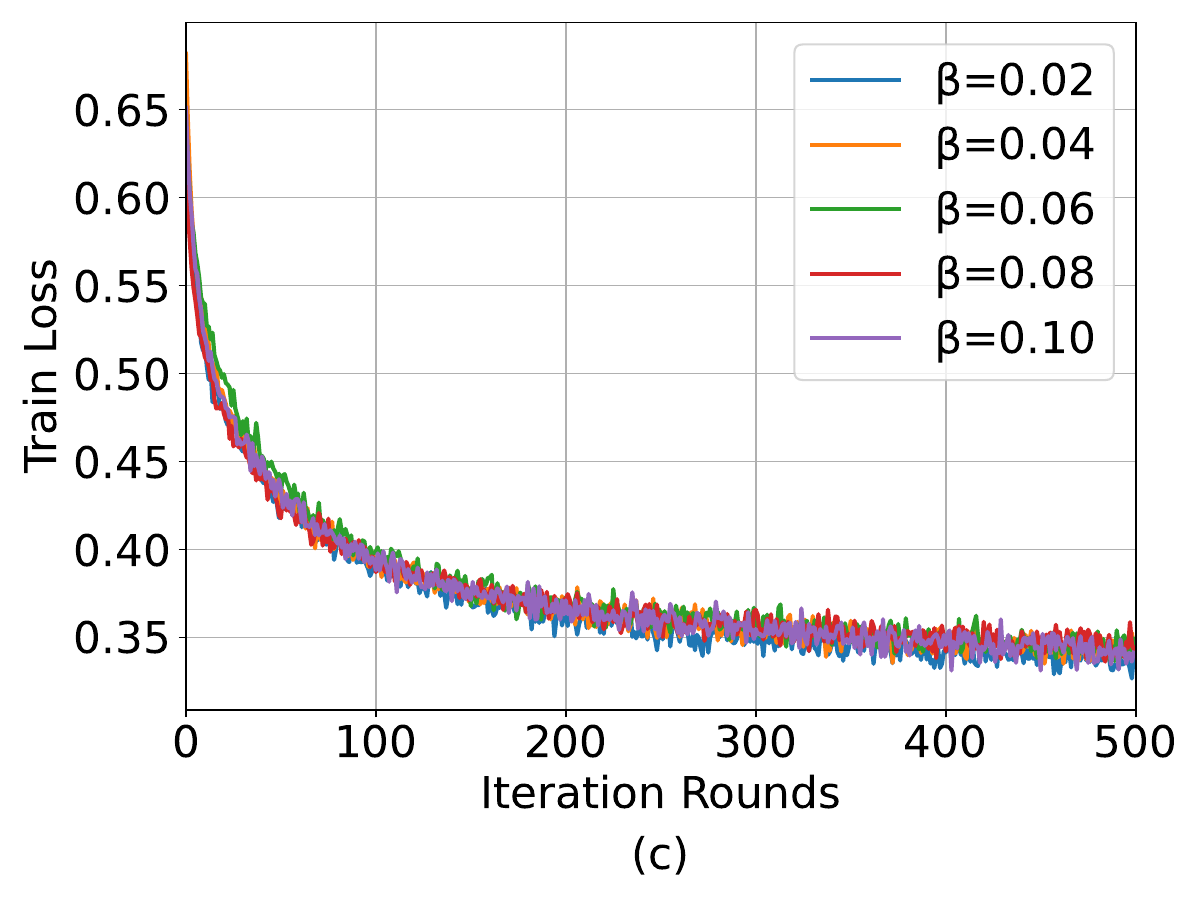}}
    \subfloat{\includegraphics[width=0.25\linewidth]{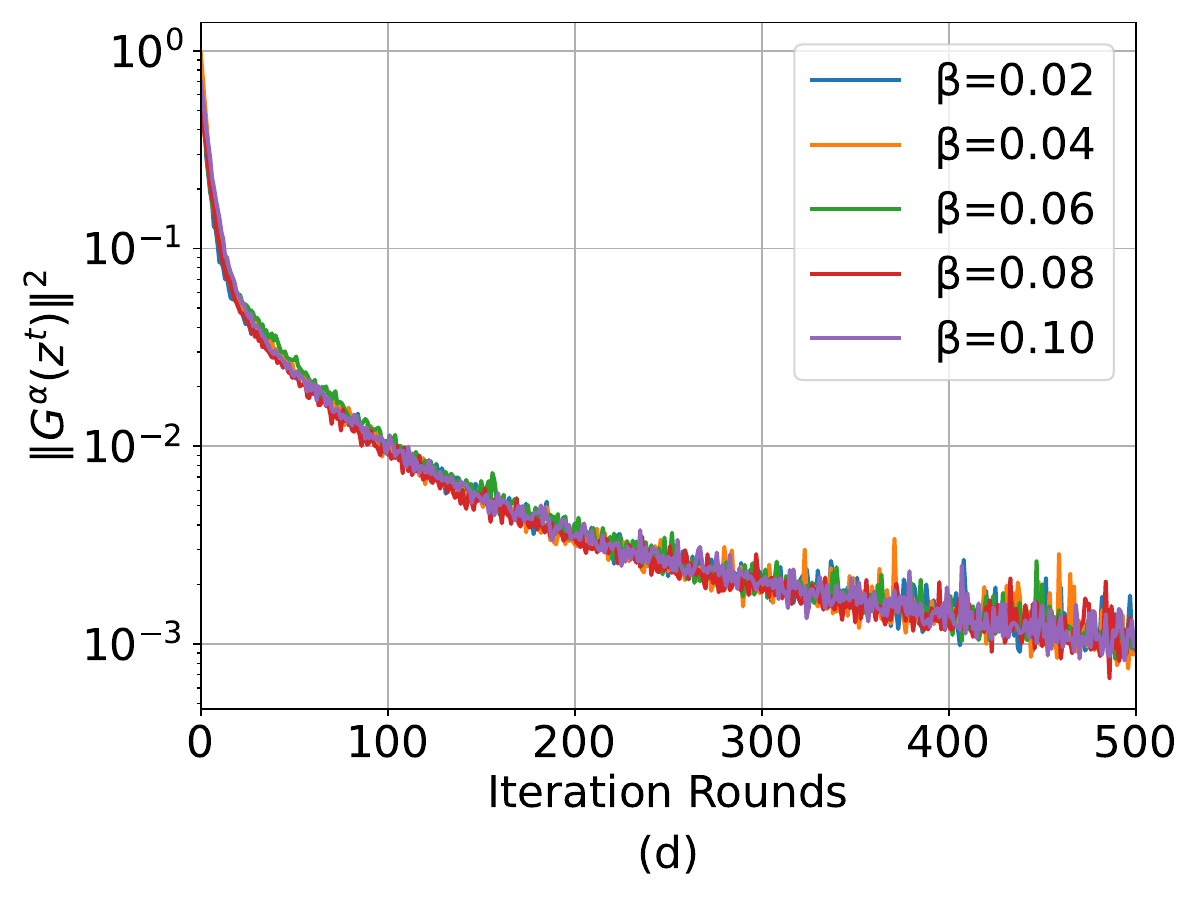}}
\vspace{-10pt}
    \caption{Training loss and proximal gradient variations of FedCanon under different global and local step sizes $\alpha$ and $\beta$: (a) and (b) use the same $\beta = 0.02$ while varying $\alpha$; (c) and (d) use the same $\alpha = 0.05$ while varying $\beta$.}\label{fig:stepsize}
\end{figure*}
\begin{figure*}[htbp]
\centering
\includegraphics[width=.32\linewidth]{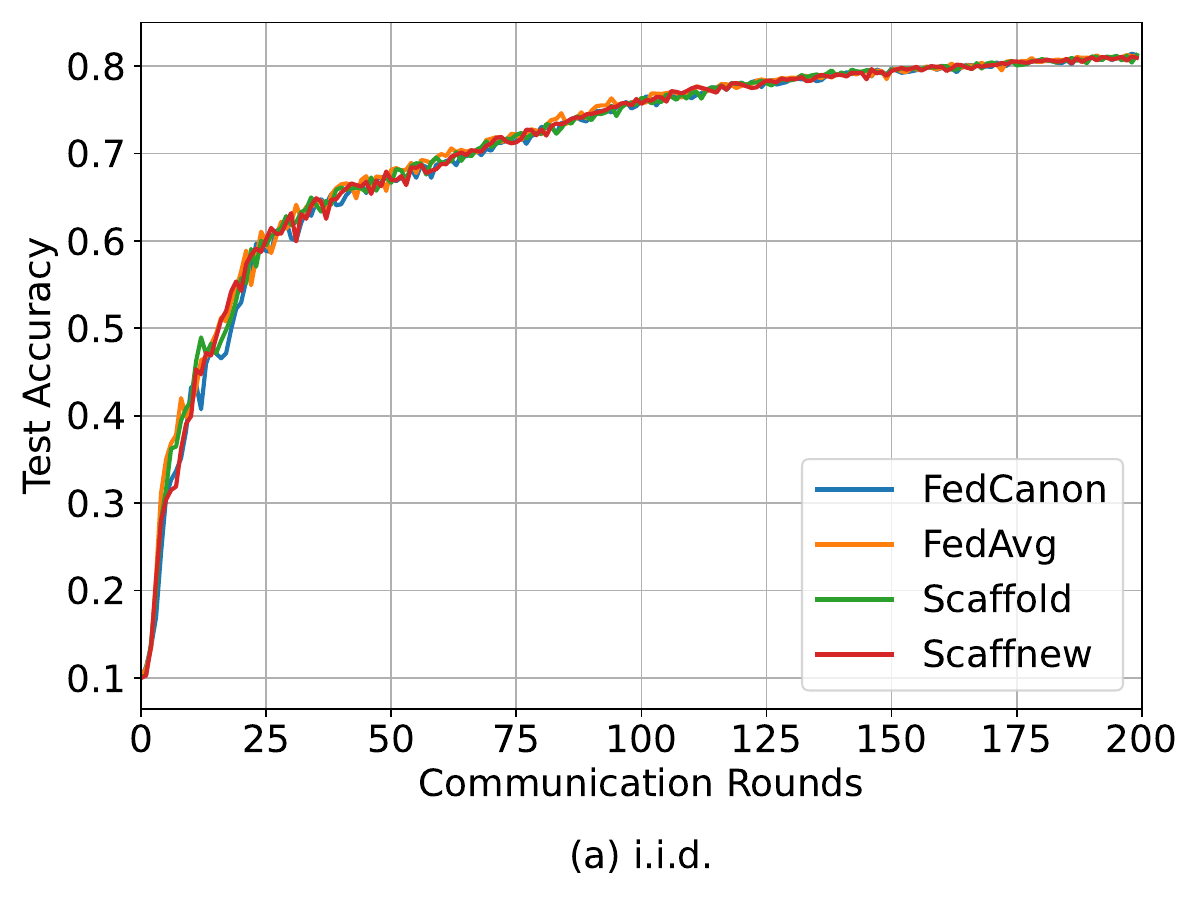}
\includegraphics[width=.32\linewidth]{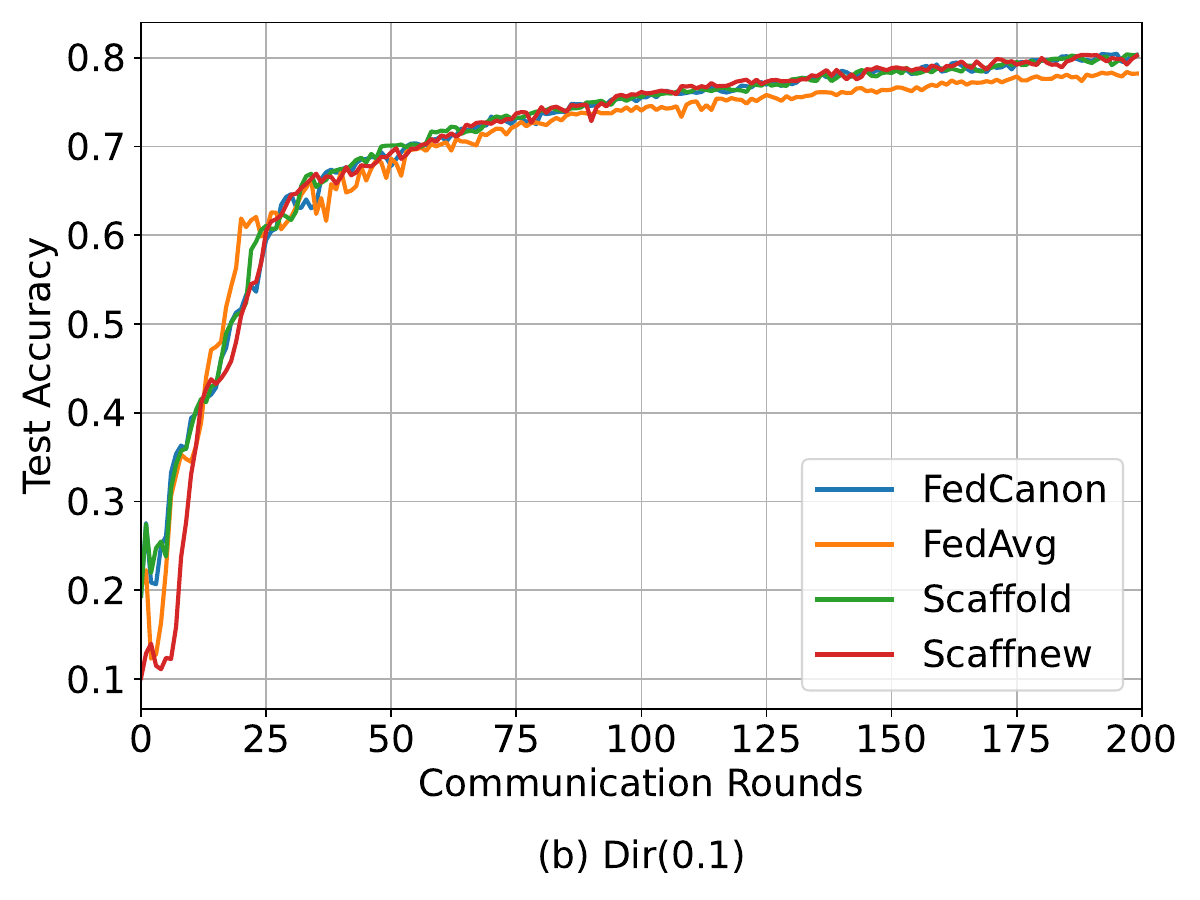}
\includegraphics[width=.32\linewidth]{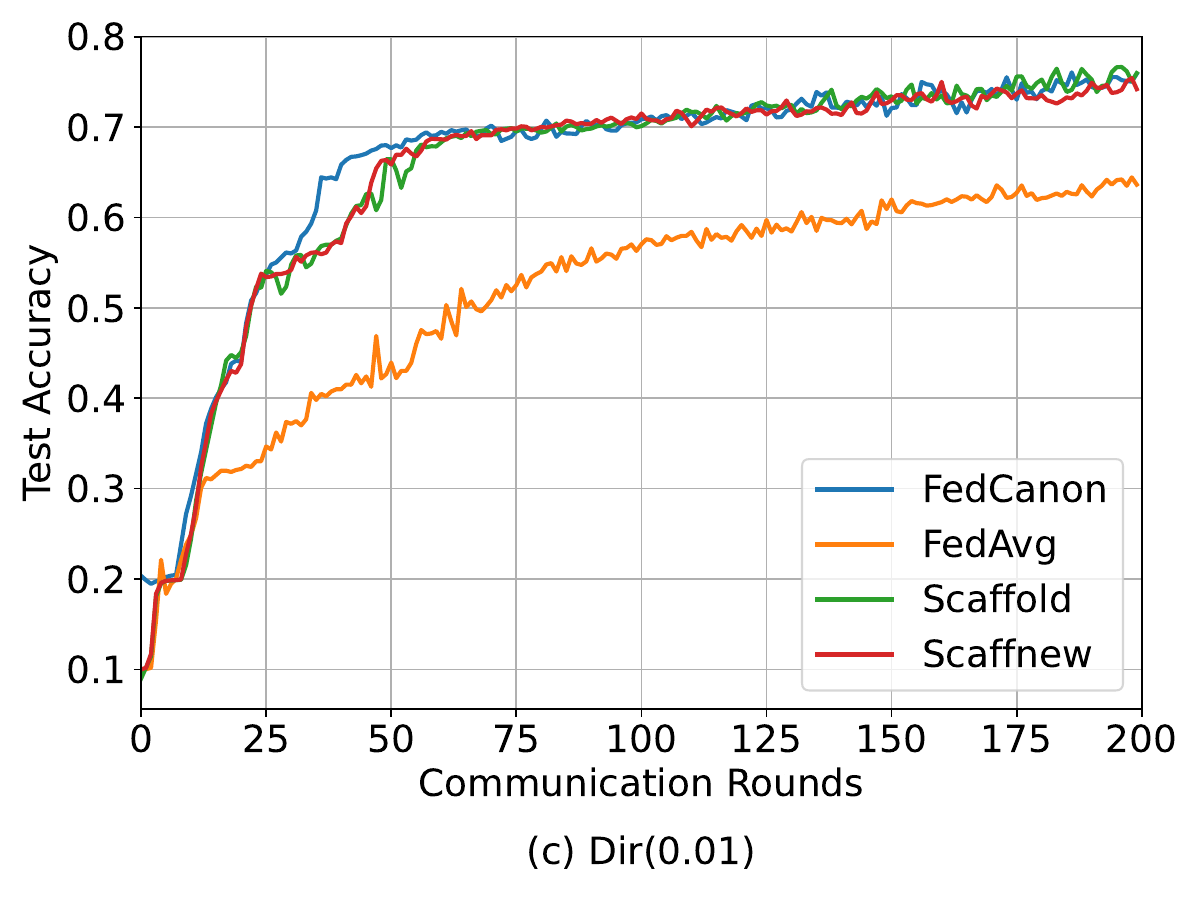}
\vspace{-10pt}
\caption{Performance of FedCanon, FedAvg, SCAFFOLD and SCAFFNEW under different levels of data heterogeneity.}\label{fig:control_variable}
\end{figure*}

Typically, the choice of step size parameters directly influences the convergence rate of FL methods. 
To investigate this effect, we uniformly distribute the A9A dataset across $10$ clients to train a simple linear model for a binary classification task, with MCP used as the regularization term.
By fixing other parameters, we plot the training loss and proximal gradient trajectories of FedCanon under various combinations of $\alpha$ and $\beta$ in Fig. \ref{fig:stepsize}. 
It can be observed that increasing $\alpha$ appropriately, while keeping $\beta$ fixed, leads to a faster decrease in training loss and proximal gradient. 
In contrast, adjusting $\beta$ while fixing $\alpha$ has a negligible influence. 
This is because the information aggregated at the server is normalized by $\beta$.

For a smooth objective problem (i.e., $h = 0$), we compare FedCanon with FedAvg, SCAFFOLD and SCAFFNEW. 
For all these algorithms, we set the same number of clients ($n=10$), learning rate, local update steps ($K=20$), and training epochs ($T=200$). 
The FMNIST dataset is used to train an MLP model under three different label distribution settings: i.i.d., weak heterogeneity ($Dir(0.1)$), and strong heterogeneity ($Dir(0.01)$).
The results are presented in Fig. \ref{fig:control_variable}. 
In the i.i.d. case (Fig. \ref{fig:control_variable} (a)), all algorithms achieve similar performance. 
Under increasing heterogeneity (Fig. \ref{fig:control_variable} (b) and (c)), FedAvg shows noticeable degradation, with test accuracy dropping by approximately $2\%$ and $10\%$, respectively, compared to the other algorithms.
In contrast, FedCanon, SCAFFOLD, and SCAFFNEW maintain comparable performance and effectively alleviate the impact of data heterogeneity.

\begin{figure*}[tb]
\centering
\includegraphics[width=.32\linewidth]{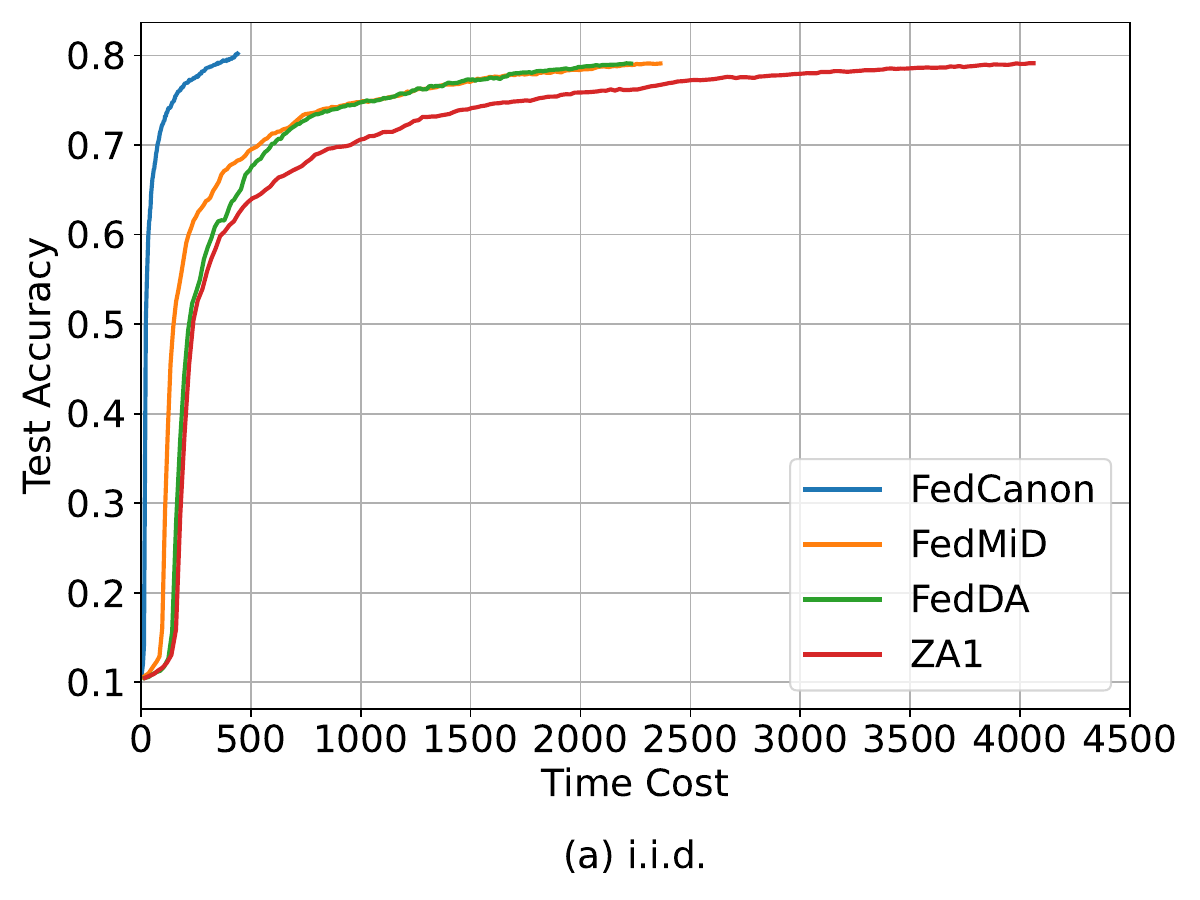}
\includegraphics[width=.32\linewidth]{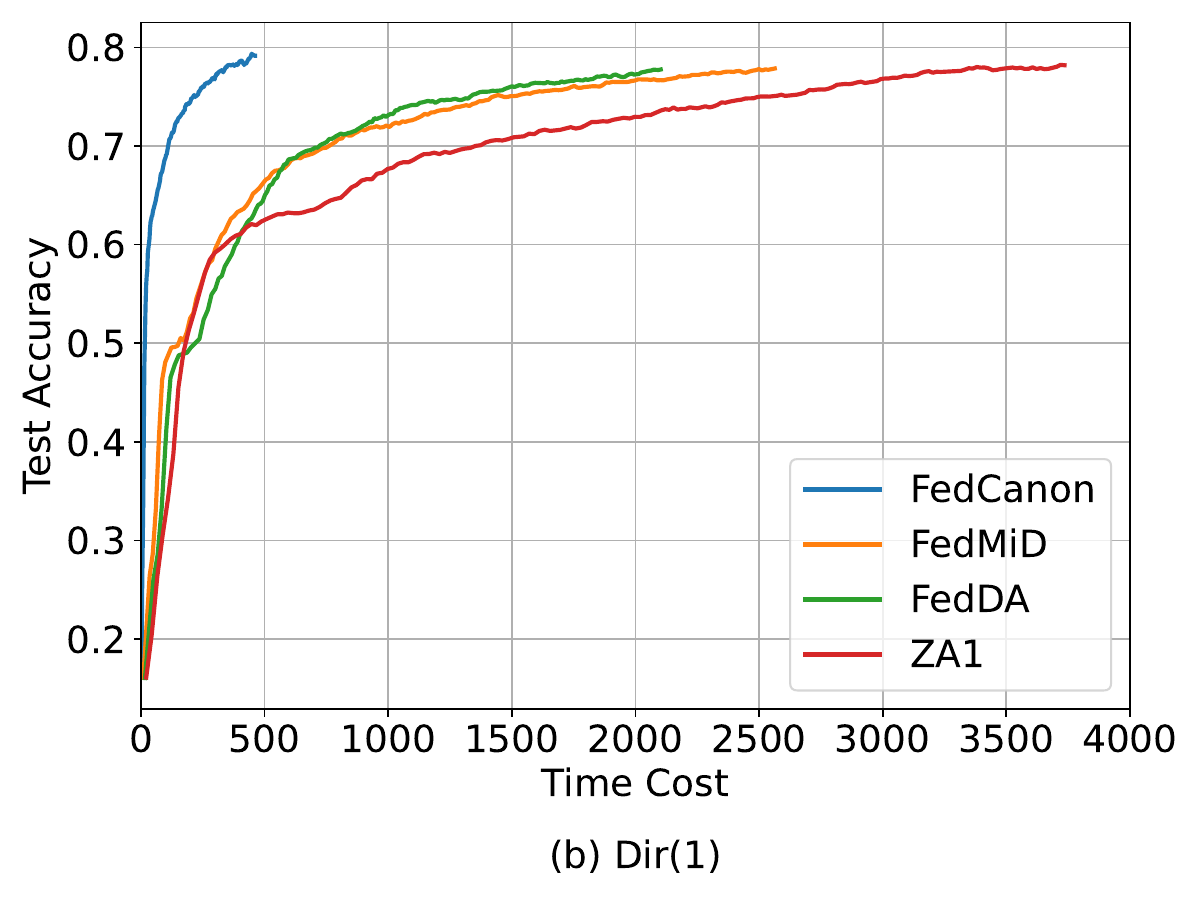}
\includegraphics[width=.32\linewidth]{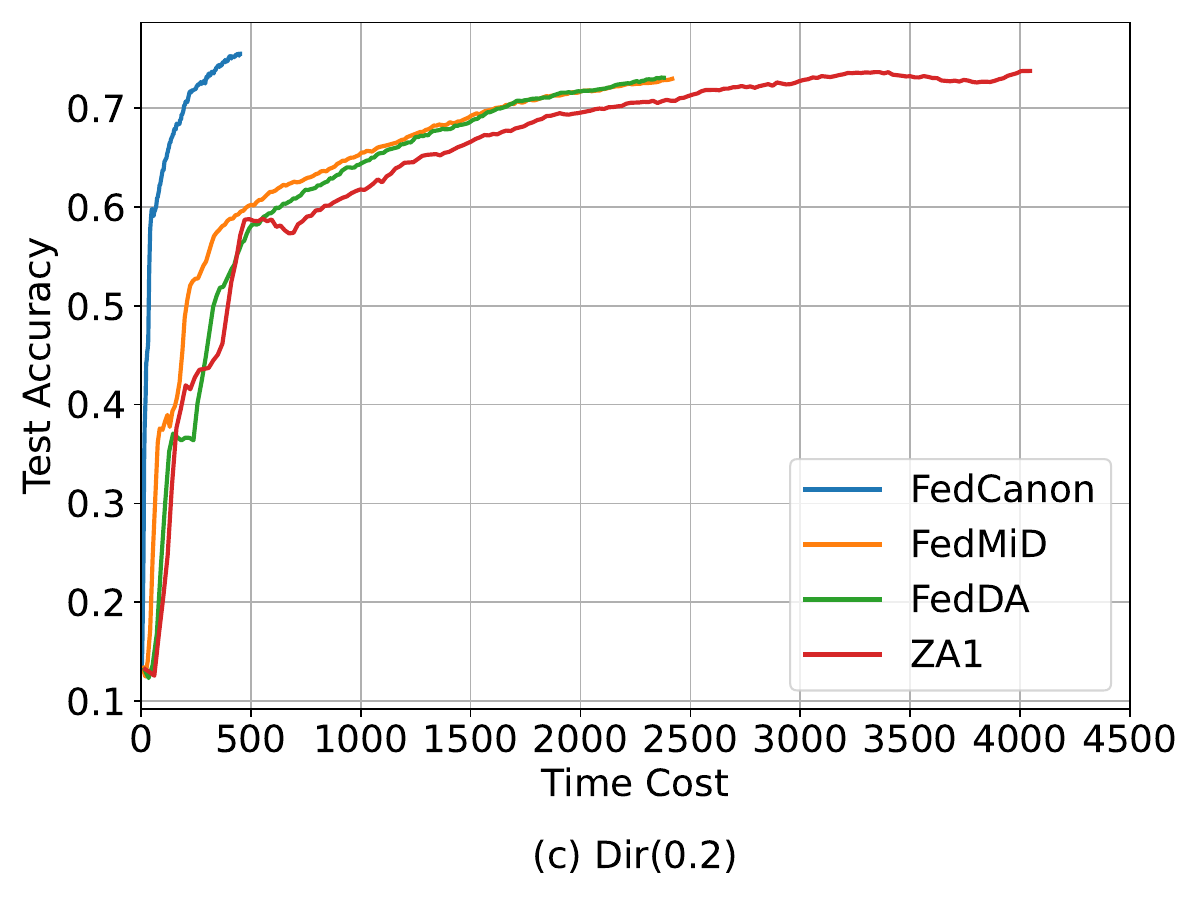}
\vspace{-10pt}
\caption{Test accuracy variations over training time (seconds) for FedCanon, FedMid, FedDA and ZA1, under different levels of data heterogeneity.}\label{fig:communication_efficient}
\end{figure*}
\begin{table}[htbp]
\renewcommand\arraystretch{1.25}
\setlength{\tabcolsep}{6pt}
\centering
\begin{threeparttable}
\caption{Total Computation Time (seconds) for Different Algorithms and Local Update Steps $K$ with Fixed Communication Rounds $T=200$.
}\label{tab:time_iid}
\begin{tabular}{c|cccc}
\toprule
 \textbf{K} & \textbf{FedCanon} & \textbf{FedMiD} & \textbf{FedDA} & \textbf{ZA1}  \\
\midrule
 \textbf{10} &  {{230}} &\makecell{{662}\\ \textcolor{blue}{ $ 2.87\times$}}  &\makecell{{804}\\ \textcolor{blue}{ $ 3.49\times$} }  & \makecell{{1106} \\ \textcolor{blue}{ $ 4.80\times$} } \\
 \hline
  \textbf{20} &\makecell{{293} \\ \textcolor{red}{ $ 1.27\times$}} &\makecell{{1132} \\ \textcolor{blue}{ $ 3.86\times$} \textcolor{red}{ $ 1.71\times$}} &\makecell{{1281} \\ \textcolor{blue}{ $ 4.37\times$} \textcolor{red}{ $ 1.59\times$}} &\makecell{{2048} \\ \textcolor{blue}{ $ 6.99\times$} \textcolor{red}{ $ 1.85\times$} } \\
   \hline
   \textbf{40} &\makecell{{417}\\ \textcolor{red}{ $ 1.42\times$}} &\makecell{{2364} \\ \textcolor{blue}{ $ 5.67\times$} \textcolor{red}{ $ 2.09\times$}} &\makecell{{2230} \\ \textcolor{blue}{ $ 5.35\times$} \textcolor{red}{ $ 1.74\times$}}  &\makecell{{4062}\\ \textcolor{blue}{ $ 9.74\times$} \textcolor{red}{ $ 1.98\times$}}  \\
    \hline
 \textbf{80} &\makecell{{719}\\  \textcolor{red}{ $ 1.72\times$}}& \makecell{{3875}\\ \textcolor{blue}{ $ 5.39\times$} \textcolor{red}{ $ 1.64\times$}}  &\makecell{{4126}\\ \textcolor{blue}{ $ 5.74\times$} \textcolor{red}{ $ 1.85\times$}}  &\makecell{{7201}\\ \textcolor{blue}{ $ 10.02\times$} \textcolor{red}{ $ 1.77\times$}}  \\
\bottomrule
\end{tabular}
\end{threeparttable}
\begin{tablenotes}[flushleft]
\item  The values in \textcolor{blue}{BLUE} indicate the time ratio of each algorithm relative to FedCanon (first column) under the same $K$ setting, while the values in \textcolor{red}{RED} represent the time ratio relative to the same algorithm in the previous row.
\end{tablenotes}
\end{table}
\begin{table*}[htbp]
\renewcommand\arraystretch{1.2}
\centering
{
\caption{Test Accuracy of FedCanon, FedDR, and FedADMM Across Different Datasets, Models and Data Distributions.
}\label{table:accuracy}
\setlength{\tabcolsep}{8pt}
\begin{tabular}{cccccc}
\toprule
\textbf{Datasets} & \textbf{Models} &\textbf{Partitions} &  \textbf{FedCanon}  & \textbf{FedDR} & \textbf{FedADMM} \\
\midrule
\multirow{4}{*}{FEMNIST} & \multirow{2}{*}{CNN} & $i.i.d.$ & $\mathbf{0.7679\pm0.0124}$  & ${0.7130\pm0.0209}$ & ${0.7136\pm0.0254}$ \\
&  & $Dir(0.2)$ & $\mathbf{0.7570\pm0.0005}$ & $0.7059\pm0.0127$ & ${0.7095\pm 0.0058}$ \\
\cline{2-6}
& \multirow{2}{*}{ResNet-18} & $i.i.d.$ & $\mathbf{0.8523 \pm0.0015 }$  & ${ 0.7983\pm0.0140 }$ & $ 0.7980\pm 0.0219$ \\
&  & $Dir(0.2)$ & $\mathbf{0.8361 \pm0.0010 }$ & $0.7899 \pm0.0033 $ & $ 0.7939\pm0.0023  $ \\
\hline
\multirow{4}{*}{CIFAR-10} & \multirow{2}{*}{CNN} & $i.i.d.$ & $\mathbf{0.6505\pm0.0066}$  & $0.5591\pm0.0065$ & $ 0.5957\pm0.0087$ \\
&  & $Dir(0.2)$ & $\mathbf{0.6333\pm0.0131}$ & $0.5015\pm0.0065$ & $0.3557\pm0.0086$ \\
\cline{2-6}
& \multirow{2}{*}{ResNet-18} & $i.i.d.$ & $\mathbf{0.7201\pm0.0052}$  & $0.6592\pm0.0167$ & $0.6182\pm0.0143$ \\
&  & $Dir(0.2)$ & $\mathbf{0.6966\pm0.0062}$ & $0.6519\pm0.0278$ & $0.5644\pm0.0359$ \\
\hline
\multirow{4}{*}{CINIC-10} & \multirow{2}{*}{CNN} & $i.i.d.$  & $\mathbf{0.4906\pm0.0007}$  & $0.4156\pm0.0247$ & $0.4310\pm0.0108$ \\
&  & $Dir(0.2)$  &$\mathbf{0.3357\pm0.0301}$  & $0.3250\pm0.0059$ & $0.1803\pm0.0123$ \\
\cline{2-6}
& \multirow{2}{*}{ResNet-18} & $i.i.d.$ & $\mathbf{0.5491\pm0.0026}$  & $0.5415\pm0.0086$ & $0.5124\pm0.0127$ \\
& & $Dir(0.2)$ & $\mathbf{0.4956\pm0.0033}$  & $0.4519\pm0.0114$ & $0.4726\pm0.0194$ \\
\hline
\multirow{2}{*}{TinyImageNet} &  \multirow{2}{*}{ResNet-18} & $i.i.d.$ & $\mathbf{0.3569\pm0.0045}$  & $0.2208\pm0.0176$ & $0.2206\pm0.0177$ \\
& & $Dir(0.2)$ & $\mathbf{0.2969\pm0.0024}$  & $0.1849\pm0.0207$ & $0.1978\pm0.0109$ \\
\bottomrule
\end{tabular}
}
\end{table*}

Next, we compare FedCanon with several existing composite FL methods, including FedMiD, FedDA and ZA1, under different data distributions (i.i.d., $Dir(1)$ and $Dir(0.2)$), using the same number of communication rounds ($T = 200$).
The FMNIST dataset is partitioned across 10 clients to train a CNN model with the SCAD penalty. 
Fig. \ref{fig:communication_efficient} illustrates the test accuracy over training time for different algorithms with a fixed number of local updates ($K=40$).
Further, TABLE \ref{tab:time_iid} provides a more detailed comparison of the training time under the i.i.d. setting for varying local update steps ($K=10, 20, 40, 80$).
As a representative example, the value in the third row and second column indicates that, when $K=40$, FedMiD consumes 5.67 times more computation than FedCanon, and 2.09 times more than when $K=20$. 
Notably, for the same number of communication rounds, FedCanon achieves superior test accuracy while requiring less training time. 
This advantage is attributed to its efficient structure and reduced local computation, underscoring its suitability for composite optimization in heterogeneous data settings.

{To provide a comprehensive performance comparison, we evaluate the test accuracy of FedCanon against the primal-dual baselines, FedDR and FedADMM. 
This evaluation spans four datasets (FEMNIST, CIFAR-10, CINIC-10 and TinyImageNet), two model architectures (CNN and ResNet-18), and both i.i.d. and non-i.i.d. ($Dir(0.2)$) partitions.
As discussed previously, a key challenge for primal-dual methods is the need to accurately solve local subproblems. 
To emulate a practical stochastic setting, we configure their clients to approximately solve these subproblems using SGD over several local updates. 
For FedCanon, we use a fixed global step size of $\alpha=0.1$ across most experiments.
The results are summarized in TABLE \ref{table:accuracy}, showing that FedCanon consistently outperforms both FedDR and FedADMM across all experimental settings. 
This outcome aligns with our analysis: by avoiding the complex local subproblems and the potential accumulation of inexactness errors inherent to primal-dual methods, FedCanon's direct, primal-only approach finds a higher-quality global model. 
}


\section{Conclusion}
In this paper, we proposed FedCanon, a novel composite FL algorithm designed for potentially non-convex problems with weakly convex, non-smooth regularization term.
The work introduces a simple yet powerful primal-only approach that simultaneously achieves computational efficiency and robustness to data heterogeneity. 
By decoupling the proximal mapping from local updates and delegating it to the server, FedCanon drastically reduces the computational burden on clients. 
This efficient design introduces an inherent structural drift, a challenge the proposed framework is intended to address. 
To handle statistical heterogeneity, FedCanon integrates control variables into local client updates. 
This entire architecture avoids solving the complex subproblems of primal-dual alternatives, offering a robust and practical solution.

The theoretical analysis provides the first rigorous convergence guarantees for such a framework under general non-convex assumptions, notably without the restrictive bounded heterogeneity condition.
This analysis establishes that FedCanon achieves sublinear convergence for general non-convex settings and linear convergence under the PL condition. 
Numerical experiments further validate its effectiveness, demonstrating that the proposed method achieves state-of-the-art accuracy while requiring significantly less computational time than existing composite FL methods, especially on large-scale, heterogeneous datasets.

Future work will focus on enhancing the robustness of FedCanon. 
A potential direction is to integrate it with advanced security mechanisms, such as privacy preservation \cite{zhang2025locally} and Byzantine fault tolerance \cite{zhang2024byzantine,dong2023byzantine}. 
This involves designing new strategies that combine FedCanon with robust aggregation rules to defend against malicious clients. 
Furthermore, analyzing the generalization error of this algorithm under such adversarial conditions \cite{ye2025generalization}, is another important avenue.



\begin{appendices}

\section{Relationship with Other Algorithms}
The equation \eqref{equ:iteration5} in FedCanon implies that the control variable $c_i$ encapsulates the previous global gradient information, which helps mitigate client drift caused by the heterogeneity of training data.
A similar role of this control variable can be observed in SCAFFOLD \cite{karimireddy2020scaffold} and SCAFFNEW \cite{mishchenko2022proxskip}.

\subsubsection{SCAFFOLD}
The control variable update in SCAFFOLD has two modes, and the main steps of using Option II in \cite[Algorithm 1]{karimireddy2020scaffold} are summarized as
\begin{subequations}\label{equ:SCAFFOLD}
    \begin{align}
    &\hat{x}_i^{t, 0} = z^{t}, \label{equ:SCAFFOLD0}\\
        &\hat{x}_i^{t, k+1} \!=\! \hat{x}_i^{t, k} \!- \!\beta\left[g_i(\hat{x}_i^{t, k}) \!+\! e^t\!-\! e_i^t \right]\!, k\!=\!0,\cdots,K\!-\!1, \label{equ:SCAFFOLD1}\\
        &\Delta_i^{t+1} =   z^t-\hat{x}_i^{t, K}, \label{equ:SCAFFOLD2}\\
        &\Delta e_i^{t+1} = \frac{1}{\beta K} \Delta_i^{t+1} - e^t, \label{equ:SCAFFOLD3}\\
        &e_i^{t+1} = e_i^{t} + \Delta e_i^{t+1}, \ e^0 = \frac{1}{N} \sum_{i\in[N]} e_i^0, \label{equ:SCAFFOLD4}\\
        &z^{t+1} = z^t - \frac{\alpha}{|\mathcal{S}|} \sum_{i \in \mathcal{S}\subseteq [N]} \Delta_i^{t+1}, \label{equ:SCAFFOLD5}\\
        &e^{t+1} = e^t + \frac{1}{N} \sum_{i \in \mathcal{S}} \Delta e_i^{t+1}, \label{equ:SCAFFOLD6}
    \end{align}
\end{subequations}
The term $e-e_i$ in \eqref{equ:SCAFFOLD} essentially serves the same role as $c_i$ in \eqref{equ:iteration1}. 
Specifically, it can be easily inferred from the local update step \eqref{equ:SCAFFOLD1} that
$$\frac{1}{\beta K}\Delta_i^{t+1}\overset{\eqref{equ:SCAFFOLD2}}{=}\frac{1}{\beta K} \big(z^t - \hat{x}_i^{t, K}\big)  = v_i^t + e^t- e_i^t ,$$
which corresponds to \eqref{equ:iteration2}.
Then, by setting $\mathcal{S} = [N]$, for the local and global control variables in \eqref{equ:SCAFFOLD4} and \eqref{equ:SCAFFOLD6}, we can infer 
\begin{align*}
&\ e_i^{t+1} = e_i^{t} + \Delta e_i^{t+1} \overset{\eqref{equ:SCAFFOLD3}}{=} e_i^{t} + \frac{1}{\beta K} \Delta_i^{t+1} - e^t = v_i^t, \\
    e^{t+1} &= e^t + \frac{1}{N} \sum_{i\in[N]}\Delta e_i^{t+1} \overset{\eqref{equ:SCAFFOLD3}}{=} e^t + \frac{1}{N} \sum_{i\in[N]} \left[\frac{1}{\beta K} \Delta_i^{t+1} \!-\! e^t\right] \\
    &= \bar{v}^t + e^t\!-\! \frac{1}{N} \sum_{i\in[N]} e_i^t = \frac{1}{N} \sum_{i\in[N]} e_i^{t+1} \!-\! \frac{1}{N} \sum_{i\in[N]} e_i^t + e^t. 
\end{align*}
Since the initial values of the control variables satisfy $e^0 = \frac{1}{N} \sum_{i\in[N]} e_i^0$, it is straightforward to obtain $e^t = \frac{1}{N} \sum_{i\in[N]} e_i^t$ by the above equation. 
Consequently, 
$$
    e^{t+1}-e_i^{t+1}=\bar{v}^t-v_i^t,\ \frac{1}{\beta NK} \sum_{i\in[N]}\Delta_i^{t+1} =\bar{v}^{t},
$$
are equivalent to \eqref{equ:iteration5} and \eqref{equ:iteration3}, respectively.

From the perspective of control variables, FedCanon can be considered as an extension of SCAFFOLD to composite problems.
In SCAFFOLD, both the server and clients must maintain and update control variables, resulting in the exchange of four variables per client per iteration.
In contrast, FedCanon eliminates server-side control variable, thereby reducing communication overhead to three variables per iteration.

\subsubsection{SCAFFNEW}
SCAFFNEW is an application of ProxSkip to FL, and its main steps are 
\begin{subequations}\label{equ:SCAFFNEW}
    \begin{align}
        &\hat{x}_i^{t+1} = {x}_i^{t} - \beta\left[g_i({x}_i^{t}) + e_i^t \right], \label{equ:SCAFFNEW1} \\
        &x_i^{t+1}=\begin{cases}
    \frac{1}{N}\sum_{i\in[N]}\hat{x}_i^{t+1}, & \text{with probability } p, \\
    \hat{x}_i^{t+1}, &  \text{with probability } 1-p,
\end{cases} \label{equ:SCAFFNEW2}\\
&e_i^{t+1} = e_i^{t} - \frac{p}{\beta}(x_i^{t+1}-\hat{x}_i^{t+1}),\ \sum_{i\in[N]} e_i^0=\mathbf{0}_d. \label{equ:SCAFFNEW3}
    \end{align}
\end{subequations}
SCAFFNEW introduces a small probability $p$, such that in each iteration, the clients skip communication with this probability and performs only local updates; otherwise, the server aggregates and averages all local models, and then broadcasts the result. 
If \eqref{equ:SCAFFNEW} is structured in a two-layer iterative form similar to that of FedCanon II, it can be written as
$$\hat{x}_i^{t,0}=\frac{1}{N}\sum_{i\in[N]}\hat{x}_i^{t-1,K_p^{t-1}},\hat{x}_i^{t,k+1} = \hat{x}_i^{t,k} - \beta\left[g_i(\hat{x}_i^{t,k}) + e_i^t \right],$$
$$e_i^{t+1} = e_i^{t} - \frac{p}{\beta}\big(\frac{1}{N}\sum_{i\in[N]}\hat{x}_i^{t,K_p^t}-\hat{x}_i^{t,K_p^t}\big),$$
where $K_p^t$ denotes the number of local updates in the $t$-th outer iteration. 
If $p$ in the second equation above is approximately replaced by $1/K_p^t$, then these two equations correspond to \eqref{equ:iteration1} and \eqref{equ:iteration5}, respectively.



\section{Technical Lemmas}

\begin{lemma}\label{lem:vector}
For any $x,y\in\mathbb{R}^{d}$, it holds that
\begin{subequations}
    \begin{align}
        &\langle x/\sqrt{\eta}, \sqrt{\eta} y \rangle \le \frac{1}{2\eta} \Vert x \Vert^2 + \frac{\eta}{2} \Vert y \Vert^2, \ \forall \eta > 0, \label{lem:vector2} \\
        &\Vert x + y \Vert^2 \le (1 + \eta) \Vert x \Vert^2 + (1 + \eta^{-1}) \Vert y \Vert^2, \ \forall \eta > 0, \label{lem:vector3}\\
        &-{\Vert x \Vert}^2 \le {\Vert x - y \Vert}^2-\frac{1}{2}{\Vert y \Vert}^2. \label{lem:vector4}
    \end{align}
\end{subequations}
\end{lemma}
\begin{lemma}\label{lem:norm}
For any $x_i\in\mathbb{R}^{d}$ and $i\in[N]$, it holds that
\begin{subequations}
    \begin{align}
        &\big\Vert \sum_{i\in[N]} x_i \big\Vert^2 \le N \sum_{i\in[N]}\left\Vert x_i \right\Vert^2, \label{lem:norm1}\\
        &\sum_{i\in[N]} \Vert x_i - \bar{x} \Vert^2 \le \sum_{i\in[N]} \Vert x_i \Vert^2, \ \bar{x} := \frac{1}{N} \sum_{i\in[N]} x_i. \label{lem:norm2}
    \end{align}
\end{subequations}
\end{lemma}
\begin{lemma}[\cite{chen2021distributed}] \label{lem:weakly_convex}
For any $x,y \in \mathbb{R}^{d}$, $1>\tau \rho \ge 0$ and $\tau > 0$, a proper, closed and $\rho$-weakly convex function $h:\mathbb{R}^d\rightarrow\mathbb{R}\cup\{+\infty\}$ satisfies 
\begin{subequations}
    \begin{align}
        &h(y) \le h(x) + \langle h'(y), y - x \rangle + \frac{\rho}{2}{\Vert y - x \Vert}^2,  \label{lem:weakly_convex1} \\
        &\left\Vert \textbf{prox}_{\tau h} \{x\} - \textbf{prox}_{\tau h} \{y\} \right\Vert \le \frac{1}{1 - \tau \rho} \Vert x-y \Vert, \label{lem:weakly_convex2}  
    \end{align}
\end{subequations}
where $h'(y) \in \partial h(y)$.
\end{lemma}
\begin{lemma}[\cite{karimireddy2020scaffold}] \label{lem:scaffold}
If the sequence $\{X_1, \dots, X_K\} \subset \mathbb{R}^d$ is Markovian, such that the conditional expectation satisfies $\mathbb{E}[X_k \mid X_{k-1}, \dots, X_1] = x_k$, and the conditional variance satisfies $\mathbb{E}[\|X_k - x_k\|^2 \mid x_k] \leq \sigma^2$, then $\{X_k - x_k\}$ forms a martingale. 
Then, we have the following bound:
\[
\mathbb{E}\big\|\sum_{k\in[K]} X_k\big\|^2 \leq 2 \mathbb{E}\big\|\sum_{k\in[K]} x_k\big\|^2 + 2K\sigma^2.
\]
\end{lemma}

\section{Proof of Lemma \ref{lem:lemma1}}\label{appendix_lemma1}

By directly substituting the definition of the proximal operator to \eqref{equ:prox_iteration}, it follows that
\begin{align*}
z^{t + 1} &= \textbf{prox}_{\alpha h} \{ z^{t} - \alpha \bar\Delta^{t+1} \} \notag \\
&= \arg\min_{z} \big\{ h(z) + \frac{1}{2 \alpha} \Vert z - (z^{t} - \alpha \bar\Delta^{t+1}) \Vert^2 \big\}.
\end{align*} 
Using the optimality condition of this minimization problem, we can derive the subgradient:
$$
-\frac{1}{\alpha}(z^{t+1}-z^t)-\bar\Delta^{t+1}=h'(z^{t+1})\in\partial h(z^{t+1}).
$$
Next, by applying \eqref{lem:weakly_convex1} in Lemma \ref{lem:weakly_convex} and substituting this subgradient, we can obtain
\begin{align}\label{equ:h_iteration}
&h(z^{t + 1}) \le h(z^t) + \langle h'(z^{t + 1}), z^{t + 1} - z^t \rangle+\frac{\rho}{2}{\Vert z^{t + 1} - z^t \Vert}^2 \notag \\
=& h(z^t) + \big\langle - \frac{1}{\alpha}(z^{t + 1} - z^{t}) - \bar\Delta^{t+1}, z^{t +1} - z^t \big\rangle \notag\\
&+\frac{\rho}{2}{\Vert z^{t + 1} - z^t \Vert}^2 \notag \\
=& h(z^t) - \left\langle \bar\Delta^{t+1}, z^{t + 1} - z^t \right\rangle - \frac{2-\alpha\rho}{2\alpha}{\Vert z^{t + 1} - z^t \Vert}^2. 
\end{align}
Furthermore, since $f$ is $L$-smooth, it directly follows from \eqref{equ:lipschitz} that
\begin{equation}\label{equ:z_lsmooth}
f({z}^{t+1}) \le f({z}^{t}) + \langle \nabla f({z}^{t}), {z}^{t+1} - {z}^{t} \rangle + \frac{L}{2} \Vert {z}^{t+1} - {z}^{t} \Vert^2.
\end{equation}
Combining \eqref{equ:h_iteration} and \eqref{equ:z_lsmooth}, then substituting \eqref{equ:prox_iteration}, we obtain
\begin{align}\label{equ:phi_iteration}
&\phi({z}^{t+1}) - \phi({z}^{t}) = f({z}^{t+1}) - f({z}^{t}) + h({z}^{t+1}) - h({z}^{t})  \notag\\
\overset{\eqref{equ:prox_iteration}}{\le}& \alpha\langle \bar\Delta^{t+1} - \nabla f({z}^{t}), G^{\alpha}(z^{t}, \bar\Delta^{t+1}) \rangle \\
&- \frac{2\alpha-(\rho+L)\alpha^2}{2} \Vert G^{\alpha}(z^{t}, \bar\Delta^{t+1}) \Vert^2 \notag \\
\overset{\eqref{lem:vector2}}{\le}& {\alpha}{\Vert\bar\Delta^{t+1} - \nabla f({z}^{t}) \Vert}^2 - \frac{3\alpha-2(\rho+L)\alpha^2}{4} \Vert G^{\alpha}(z^{t}, \bar\Delta^{t+1}) \Vert^2, \notag
\end{align}
where $\eta=\frac{1}{2}$ is set in \eqref{lem:vector2}.
Next, if $1-\alpha\rho > 0$ holds, we can deduce the following relation between $G^{\alpha}(z^{t}, \bar\Delta^{t+1})$ and the proximal gradient $G^{\alpha}(z^{t})$:
\begin{align}\label{equ:prox_gradient_iteration}
&-\Vert G^{\alpha}(z^{t}, \bar\Delta^{t+1}) \Vert^2 \\
\overset{\eqref{lem:vector4}}{\le}& \Vert G^{\alpha}(z^{t}, \bar\Delta^{t+1}) - G^{\alpha}(z^{t}) \Vert^2- \frac{1}{2}\Vert G^{\alpha} (z^{t})\Vert^2 \notag \\
=&\big\Vert  \frac{1}{\alpha}\textbf{prox}_{\alpha h} \{ z^{t} - \alpha \bar\Delta^{t+1} \} - \frac{1}{\alpha} \textbf{prox}_{\alpha h} \{ z^{t} - \alpha \nabla f(z^{t}) \} \big\Vert^2 \notag \\
&- \frac{1}{2}\Vert G^{\alpha} (z^{t})\Vert^2 \notag \\
\overset{\eqref{lem:weakly_convex2}}{\le}& \frac{1}{(1-\alpha\rho)^2}\Vert \bar\Delta^{t+1} - \nabla f(z^{t}) \Vert^2- \frac{1}{2}\Vert G^{\alpha} (z^{t})\Vert^2. \notag
\end{align}
Similarly, the relationship between the proximal gradient $G^{\alpha}(z^{t})$ and the gradient of loss function $\nabla f(z^t)$ is given by
\begin{align}\label{equ:prox_gradient_iteration2}
&-\Vert G^{\alpha}(z^{t}) \Vert^2 \overset{\eqref{lem:vector4}}{\le} \Vert G^{\alpha}(z^{t}) - \nabla f(z^{t}) \Vert^2- \frac{1}{2}\Vert \nabla f(z^{t})\Vert^2 \notag \\
{\le}& \big\Vert \frac{1}{\alpha} [z^{t}\!-\!\alpha \nabla f(z^{t})\!-\!\underbrace{\textbf{prox}_{\alpha h} \{ z^{t} \!- \!\alpha \nabla f(z^{t}) \}}_{z^{+}}] \big\Vert^2\!-\! \frac{1}{2}\Vert \nabla f(z^{t})\Vert^2 \notag \\
{\le}& B_h- \frac{1}{2}\Vert \nabla f(z^{t})\Vert^2,
\end{align}
where the final inequality follows from the optimality condition of the proximal operator 
$$
-\frac{1}{\alpha}(z^{+}-z^t)-\nabla f(z^{t})=h'(z^{+})\in\partial h(z^{+}),
$$
and the assumption that the subgradients of the regularization term $h$ are bounded.
Then, plugging \eqref{equ:prox_gradient_iteration} and \eqref{equ:prox_gradient_iteration2} into \eqref{equ:phi_iteration}, it follows that
\begin{align}\label{equ:phi_iteration2}
&\phi({z}^{t+1}) - \phi({z}^{t}) \notag \\
\le& \alpha {\Vert\bar\Delta^{t+1} - \nabla f({z}^{t}) \Vert}^2 - \frac{\alpha-2(\rho+L)\alpha^2}{4} \Vert G^{\alpha}(z^{t}, \bar\Delta^{t+1}) \Vert^2 \notag\\
&- \frac{\alpha}{2} \Vert G^{\alpha}(z^{t}, \bar\Delta^{t+1}) \Vert^2  \notag\\
\overset{\eqref{equ:prox_gradient_iteration}}{\le} &\frac{(2\!+\!\delta)\alpha}{2} {\Vert\bar\Delta^{t+1}\!-\! \nabla f({z}^{t}) \Vert}^2\!-\!\frac{\alpha\!-\!2(\rho\!+\!L)\alpha^2}{4} \Vert G^{\alpha}(z^{t}, \bar\Delta^{t+1}) \Vert^2 \notag \\
&-\frac{\alpha}{4}\Vert G^{\alpha}(z^{t}) \Vert^2  \notag \\
\overset{\eqref{equ:prox_gradient_iteration2}}{\le} &\frac{(2\!+\!\delta)\alpha}{2} {\Vert\bar\Delta^{t+1}\!-\! \nabla f({z}^{t}) \Vert}^2\!-\!\frac{\alpha\!-\!2(\rho\!+\!L)\alpha^2}{4}  \Vert G^{\alpha}(z^{t}, \bar\Delta^{t+1}) \Vert^2 \notag \\
&-\frac{\alpha}{8}\Vert G^{\alpha}(z^{t}) \Vert^2-\frac{\alpha}{16}\Vert \nabla f(z^{t})\Vert^2+\frac{\alpha B_h}{8}. 
\end{align}
    Next, we focus on the expectation of the first term on the right-hand side of \eqref{equ:phi_iteration2}:
\begin{align}\label{equ:gradient_error2}
&\mathbb{E}\Vert \bar\Delta^{t+1} - \nabla f({z}^{t}) \Vert^2\overset{\eqref{equ:iteration3}}{=}\mathbb{E}\Vert \bar v^{t} - \nabla f({z}^{t}) \Vert^2 \notag\\
=&\mathbb{E}\left\Vert \frac{1}{NK}\sum_{i=1}^{N} \sum_{k=0}^{K-1} g_i(\hat{x}_i^{t, k}) - \frac{1}{N}\sum_{i=1}^{N}\nabla f_i(\hat{x}_i^{t, 0})  \right\Vert^2 \notag\\
\overset{\eqref{lem:norm1}}{\le}&\frac{1}{NK^2}\sum_{i=1}^{N}\mathbb{E}\left\Vert   \sum_{k=0}^{K-1} \left[g_i(\hat{x}_i^{t, k}) - \nabla f_i(\hat{x}_i^{t, 0})\right] \right\Vert^2 \notag \\
\overset{\text{Lemma } \ref{lem:scaffold}}{\le}&   \frac{2}{NK^2}\sum_{i=1}^{N} \mathbb{E}\left\Vert   \sum_{k=0}^{K-1} \left[\nabla f_i(\hat{x}_i^{t, k}) -  \nabla f_i(\hat{x}_i^{t, 0})\right] \right\Vert^2+ \frac{2\sigma^2}{BK} \notag\\
\overset{\eqref{equ:lipschitz2},\eqref{lem:norm1}}{\le}&\frac{2L^2}{NK} \sum_{i=1}^{N} \sum_{k=0}^{K-1} \mathbb{E}\Vert \hat{x}_i^{t, k} - \hat{x}_i^{t, 0} \Vert^2+\frac{2\sigma^2}{BK},
\end{align}
Moreover, for $\mathbb{E}{\Vert \hat{x}_i^{t, k} - \hat{x}_i^{t, 0} \Vert^2}$ in \eqref{equ:gradient_error2}, we can derive the following results:
\begin{align*}
&\mathbb{E}{\Vert \hat{x}_i^{t, k} - \hat{x}_i^{t, 0} \Vert^2} \overset{\eqref{equ:iteration1}}{=} \mathbb{E}{\Vert \hat{x}_i^{t, k-1} - \hat{x}_i^{t, 0} - \beta[g_i(\hat{x}_i^{t, k-1}) + c_i^t] \Vert^2} \\
\overset{\eqref{lem:vector3}}{\le}& \frac{K}{K-1} \mathbb{E}{\Vert \hat{x}_i^{t, k-1}-\hat{x}_i^{t, 0} \Vert^2} + \beta^2K \mathbb{E}{\Vert g_i(\hat{x}_i^{t, k-1}) + c_i^t \Vert^2} \\
\overset{\eqref{equ:minibatch}}{\le}& \frac{K}{K-1} \mathbb{E}{\Vert \hat{x}_i^{t, k-1}-\hat{x}_i^{t, 0} \Vert^2}  \\
&+ \beta^2K \mathbb{E}{\Vert \nabla f_i(\hat{x}_i^{t, k-1})+ c_i^t \Vert^2}+ \frac{\beta^2K\sigma^2}{B} \\
\overset{\eqref{lem:vector3}}{\le}& \frac{K}{K-1}\mathbb{E}{\Vert \hat{x}_i^{t, k-1}-\hat{x}_i^{t, 0} \Vert^2}+ 2\beta^2K \mathbb{E}{\Vert \nabla f_i(\hat{x}_i^{t, 0}) + c_i^t \Vert^2} \\
&+2\beta^2K \mathbb{E}{\Vert \nabla f_i(\hat{x}_i^{t, k-1})-\nabla f_i(\hat{x}_i^{t, 0}) \Vert^2}+ \frac{\beta^2K\sigma^2}{B} \\
\overset{\eqref{equ:lipschitz2}}{\le}& \left(\frac{K}{K-1}+ 2\beta^2KL^2\right) \mathbb{E}{\Vert \hat{x}_i^{t, k-1} - \hat{x}_i^{t, 0} \Vert^2}  \\
&+ 2\beta^2K \mathbb{E}{\Vert \nabla f_i(\hat{x}_i^{t, 0}) + c_i^t \Vert^2}+ \frac{\beta^2K\sigma^2}{B} \\
\overset{(a)}{\le}& \beta^2K\sum_{r=0}^{k-1}\!\left(\!\frac{K}{K\!-\!1}\!+\! 2\beta^2KL^2\!\right)^r\!\left[\!  2\mathbb{E}{\Vert \nabla f_i(\hat{x}_i^{t, 0})\!+\!c_i^t\Vert}^2\!+\!\frac{\sigma^2}{B}\!\right] \\
\le& \beta^2K^2\!\left(\!\frac{K}{K\!-\!1}\!+\! 2\beta^2KL^2\!\right)^{K\!-\!1}\!\left[2\mathbb{E}{\Vert \nabla f_i(\hat{x}_i^{t, 0})\!+\!c_i^t\Vert}^2\!+\!\frac{\sigma^2}{B}\!\right],
\end{align*}
where $(a)$ is derived by iteratively substituting \eqref{equ:iteration1} back to $\hat{x}_i^{t, 0} = z^{t}$.
Furthermore, if $2\beta^2KL^2 \le \frac{1}{12(K-1)}$, by leveraging the property $\lim_{x \to \infty} (1 + \frac{1}{x} + \frac{1}{12x})^x = e^{1 + \frac{1}{12}} < 3$, we can derive another upper bound:
\begin{align}\label{equ:k_update_error2}
&\sum_{i=1}^{N} \sum_{k=0}^{K-1} \mathbb{E}{\Vert \hat{x}_i^{t, k} - \hat{x}_i^{t, 0} \Vert^2}   \\
\le&6\beta^2K^2\sum_{i=1}^{N} \sum_{k=0}^{K-1}\mathbb{E}{\Vert \nabla f_i(\hat{x}_i^{t, 0})+c_i^t\Vert}^2+\frac{3\beta^2NK^3\sigma^2}{B}  \notag \\
\overset{\eqref{equ:iteration5}}{=}&6\beta^2K^3 \sum_{i=1}^{N} \mathbb{E}{\Vert \nabla f_i(\hat{x}_i^{t, 0}) - v_i^{t-1} + \bar{v}^{t-1}\Vert^2}+\frac{3\beta^2NK^3\sigma^2}{B} \notag \\
=&6\beta^2K^3 \sum_{i=1}^{N} \mathbb{E}{\Vert \nabla f_i(\hat{x}_i^{t, 0}) \!-\! v_i^{t-1} \!-\! \nabla f({z}^{t}) \!+\! \bar{v}^{t-1} \!+\! \nabla f({z}^{t}) \Vert^2} \notag \\
&+\frac{3\beta^2NK^3\sigma^2}{B} \notag \\
\overset{\eqref{lem:vector3}}{\le}&  12\beta^2NK^3 \mathcal{E}^t+ 12\beta^2NK^3 \mathbb{E}{\Vert \nabla f({z}^{t}) \Vert^2}+ \frac{3\beta^2NK^3\sigma^2}{B}. \notag
\end{align}
Then, by substituting \eqref{equ:gradient_error2}, \eqref{equ:k_update_error2} into \eqref{equ:phi_iteration2}, the desired equation \eqref{equ:phi_iteration3} can be obtained:
\begin{align*}
&\mathbb{E}[\phi ({z}^{t+1}) - \phi({z}^{t})]  \\
\le &\frac{(2+\delta)\alpha}{2} \mathbb{E}{\Vert\bar\Delta^{t+1} - \nabla f({z}^{t}) \Vert}^2 -\frac{\alpha}{8}\mathbb{E}\Vert G^{\alpha}(z^{t}) \Vert^2  \\
&\!-\!\frac{\alpha\!-\!2(\rho\!+\!L)\alpha^2}{4} \mathbb{E}\Vert G^{\alpha}(z^{t}, \bar\Delta^{t+1}) \Vert^2\!-\!\frac{\alpha}{16}\mathbb{E}\Vert \nabla f(z^{t})\Vert^2\!+\!\frac{\alpha B_h}{8} \notag \\
\overset{\eqref{equ:gradient_error2}}{\le} &\frac{\alpha(2 + \delta)L^2}{NK}   \sum_{i=1}^{N} \sum_{k=0}^{K-1} \mathbb{E}\Vert  \hat{x}_i^{t, k} - \hat{x}_i^{t, 0} \Vert^2   -\frac{\alpha}{8}\mathbb{E}\Vert G^{\alpha}(z^{t}) \Vert^2 \notag \\
&- \frac{\alpha-2(\rho+L)\alpha^2}{4} \mathbb{E}\Vert G^{\alpha}(z^{t}, \bar\Delta^{t+1}) \Vert^2-\frac{\alpha}{16}\mathbb{E}\Vert \nabla f(z^{t})\Vert^2 \notag \\
&+ \frac{\alpha(2 + \delta)\sigma^2}{KB}+\frac{\alpha B_h}{8} \notag \\
\overset{\eqref{equ:k_update_error2}}{\le} 
&- \frac{\alpha-2(\rho+L)\alpha^2}{4} \mathbb{E}\Vert G^{\alpha}(z^{t}, \bar\Delta^{t+1}) \Vert^2 -\frac{\alpha}{8}\mathbb{E}\Vert G^{\alpha}(z^{t}) \Vert^2 \\
&-\left[\frac{\alpha}{16}-{12(2 + \delta)\alpha\beta^2K^2L^2}\right]\mathbb{E}\Vert \nabla f(z^{t})\Vert^2+\frac{\alpha B_h}{8} \\
&+{12(2 + \delta)\alpha\beta^2K^2L^2}   \mathcal{E}^t + \frac{\alpha(2 + \delta)(1+3\beta^2K^3L^2)\sigma^2}{KB}. \notag 
\end{align*}

\section{Proof of Lemma \ref{lem:lemma2}} \label{appendix_lemma2}
The detailed derivation of \eqref{equ:gradient_track} is given as follows:
\begin{align*}
&\mathcal{E}^{t+1}\overset{\eqref{lem:norm2}}{\le} \frac{1}{N}\sum_{i=1}^{N} \mathbb{E}{\Vert \nabla f_i(\hat{x}_i^{t+1, 0}) - v_i^{t} \Vert^2} \notag \\
\overset{\eqref{lem:vector3}}{\le}&\frac{2}{N}\sum_{i=1}^{N} \mathbb{E}{\Vert \nabla f_i(\hat{x}_i^{t, 0}) - v_i^{t} \Vert^2} \\
&+\frac{2}{N}\sum_{i=1}^{N} \mathbb{E}{\Vert \nabla f_i(\hat{x}_i^{t+1, 0})- \nabla f_i(\hat{x}_i^{t, 0}) \Vert^2} \notag \\
\overset{\eqref{equ:lipschitz2}}{\le}&\frac{2}{N}\sum_{i=1}^{N} \mathbb{E}{\Vert \nabla f_i(\hat{x}_i^{t, 0}) - v_i^{t} \Vert^2} + 2L^2 \mathbb{E}\Vert {z}^{t+1} - {z}^{t} \Vert^2 \\
\overset{\eqref{equ:prox_iteration}}{=}& \frac{2}{N}\sum_{i=1}^{N} \mathbb{E}{\Vert \nabla f_i(\hat{x}_i^{t, 0}) - \frac{1}{K} \sum_{k=0}^{K-1} g_i(\hat{x}_i^{t, k}) \Vert^2} \\
&+ 2\alpha^2L^2 \mathbb{E}\Vert G^{\alpha}(z^{t}, \bar\Delta^{t+1}) \Vert^2  \notag \\
\overset{\text{Lemma } \ref{lem:scaffold}}{\le}& \frac{4}{N}\sum_{i=1}^{N} \mathbb{E}{\Vert \nabla f_i(\hat{x}_i^{t, 0}) - \frac{1}{K} \sum_{k=0}^{K-1} \nabla f_i(\hat{x}_i^{t,k}) \Vert^2} \\
&+ 2\alpha^2L^2 \mathbb{E}\Vert G^{\alpha}(z^{t}, \bar\Delta^{t+1}) \Vert^2+ \frac{4\sigma^2}{BK} \notag \\
\overset{\eqref{equ:lipschitz2}}{\le}&\frac{4L^2}{NK} \sum_{i=1}^{N}\sum_{k=0}^{K-1}\mathbb{E}\Vert \hat{x}_i^{t,k}-\hat{x}_i^{t, 0} \Vert^2 \\
&+ 2\alpha^2L^2 \mathbb{E}\Vert G^{\alpha}(z^{t}, \bar\Delta^{t+1}) \Vert^2+ \frac{4\sigma^2}{BK}  \notag \\
\overset{\eqref{equ:k_update_error2}}{\le}& {48\beta^2K^2L^2}\mathcal{E}^t+ 48\beta^2K^2L^2  \mathbb{E}{\Vert \nabla f({z}^{t}) \Vert^2} \\
&+2\alpha^2L^2 \mathbb{E}\Vert G^{\alpha}(z^{t}, \bar\Delta^{t+1}) \Vert^2 + \frac{4(1+3\beta^2K^3L^2)\sigma^2}{BK}.
\end{align*}

\section{Proof of Theorem \ref{thm:convergence}} \label{appendix_theorem1}

By combining \eqref{equ:phi_iteration3} and \eqref{equ:gradient_track}, we can construct the following inequality:
\begin{align}\label{equ:lyapunov}
&\mathbb{E}[\phi ({z}^{t+1}) - \phi({z}^{t})]+\alpha\left(\mathcal{E}^{t+1}-\mathcal{E}^{t}\right)  \\
\le& -\frac{\alpha}{8}\mathbb{E}\Vert G^{\alpha}(z^{t}) \Vert^2- \underbrace{\left[\alpha - {12(6+ \delta)\alpha \beta^2K^2L^2} \right]}_{\ge 0} \mathcal{E}^{t} \notag \\
&  - \underbrace{\left[\frac{\alpha-2(\rho+L)\alpha^2}{4}-2\alpha^3L^2\right]}_{\ge 0} \mathbb{E}\Vert G^{\alpha}(z^{t}, \bar\Delta^{t+1}) \Vert^2 \notag \\
& - \underbrace{\left[\frac{\alpha}{16}-{12(6+ \delta)\alpha \beta^2K^2L^2}\right]}_{\ge 0} \mathbb{E}\Vert \nabla f(z^{t})\Vert^2 \notag \\
&+ \frac{\alpha(6+\delta)(1+3\beta^2K^3L^2)\sigma^2}{BK}+\frac{\alpha B_h}{8}. \notag
\end{align}
To satisfy the requirements for convergence analysis, we need to ensure that the coefficients of the second and third terms on the right-hand side of \eqref{equ:lyapunov} are negative.
Therefore, we can determine a feasible range for $\alpha$ and $\beta$:
\begin{subequations}\label{equ:lyapunov_condition2}
\begin{align}
&\alpha(\rho+L)+4\alpha^2L^2 \le \frac{1}{2},  \label{equ:lyapunov_condition2_3} \\
&\beta^2 \le \frac{1}{192(6+\delta)K^2L^2}. \label{equ:lyapunov_condition2_2} 
\end{align}
\end{subequations}
If the above conditions are satisfied, we can rearrange \eqref{equ:lyapunov} to obtain
\begin{align*}
&\mathbb{E}\Vert G^{\alpha}(z^{t}) \Vert^2   \\
\le& \frac{8}{\alpha}\mathbb{E}[\phi ({z}^{t})-\phi({z}^{t+1})] + {8}\left[\mathcal{E}^t-\mathcal{E}^{t+1}\right] \\
&+ \frac{8(6+\delta)(1+3\beta^2K^3L^2)\sigma^2}{BK}+B_h \\
\overset{\eqref{equ:lyapunov_condition2}}{\le}&\frac{8}{\alpha}\mathbb{E}[\phi ({z}^{t})\!-\!\phi({z}^{t+1})] + {8}\left[\mathcal{E}^t-\mathcal{E}^{t+1}\right]+ \frac{50\sigma^2}{BK}+ \frac{\sigma^2}{8B}+B_h,
\end{align*}
By summing and telescoping the above inequality from $t=0$ to $T-1$, and taking the average, we can derive  \eqref{equ:convergence}.


\section{Proof of Theorem \ref{thm:pl_convergence}}\label{appendix_theorem2}
If the objective function satisfies the more general PL condition described in Assumption \ref{ass:pl_condition}, then 
$$
\Vert G^{\alpha}(z^{t}) \Vert^2\ge 2\mu [\phi({z}^{t})-\phi^*]
$$
naturally follows from \eqref{equ:pl_defn}. 
Subsequently, substituting it directly into \eqref{equ:phi_iteration3} yields the following inequality:
\begin{align}\label{equ:pl_iteration}
&\mathbb{E}\left[\phi ({z}^{t+1})-\phi^*\right] \\
{\le}&\left(1\!-\!\frac{\alpha\mu}{4}\right)\mathbb{E}\left[\phi({z}^{t})\!-\!\phi^* \right] - \frac{\alpha\!-\!2(\rho\!+\!L)\alpha^2}{4} \mathbb{E}\Vert G^{\alpha}(z^{t}, \bar\Delta^{t+1}) \Vert^2 \notag\\
&-\frac{\alpha}{16}\mathbb{E}\Vert \nabla f(z^{t})\Vert^2 +{12(2 + \delta)\alpha \beta^2K^2L^2}   [\mathcal{E}^t+\mathbb{E}\Vert \nabla f(z^{t})\Vert^2] \notag  \\
&+\frac{\alpha(2 + \delta)(1+3\beta^2NL^2K^3)\sigma^2}{BK}+\frac{\alpha B_h}{8}.  \notag
\end{align}
Using a similar approach to constructing \eqref{equ:lyapunov}, we combine \eqref{equ:gradient_track} and \eqref{equ:pl_iteration},  which leads to
\begin{align}\label{equ:pl_iteration2}
&\mathbb{E}\left[\phi ({z}^{t+1})-\phi^*\right] +\alpha\mathcal{E}^{t+1} \notag \\
\le&  \underbrace{\left(1-\frac{\alpha\mu}{4}\right)}_{\in(0,1)}\mathbb{E}\left[\phi({z}^{t})-\phi^* \right]+ \alpha\underbrace{\left[{12(6 + \delta)\beta^2K^2L^2}\right]}_{\le \left(1-\frac{\alpha\mu}{4}\right)} \mathcal{E}^{t} \notag \\
& - \underbrace{\left[\frac{\alpha-2(\rho+L)\alpha^2}{4}-2\alpha^3L^2\right]}_{\ge 0} \mathbb{E}\Vert G^{\alpha}(z^{t}, \bar\Delta^{t+1}) \Vert^2 \notag \\
& - \underbrace{\left[\frac{\alpha}{16}-{12\alpha(6 + \delta)\beta^2K^2L^2}\right]}_{\ge 0} \mathbb{E}\Vert \nabla f(z^{t})\Vert^2 \notag \\
&+ \frac{\alpha(6+\delta)(1+3K^3L^2\beta^2)\sigma^2}{BK}+\frac{\alpha B_h}{8}.
\end{align}
Similar to \eqref{equ:lyapunov}, we need to ensure that the coefficients on the right-hand side of \eqref{equ:pl_iteration2} satisfy the above conditions. 
Therefore, the range of values for $\alpha$ and $\beta$ is expressed as
\begin{subequations}\label{equ:pl_condition}
\begin{align}
&\alpha(\rho+L)+4\alpha^2L^2 \le \min\left\{\frac{1}{2}, \frac{4\mu(\rho+L)+64L^2}{\mu^2}\right\},  \label{equ:pl_condition_1} \\
&{12(6+\delta)\beta^2K^2L^2}\le \min\left\{\frac{1}{16},1-\frac{\alpha\mu}{4}\right\}. 
\end{align}
\end{subequations}
Then, we can simplify \eqref{equ:pl_iteration2} to the following form:
\begin{align*}
&\mathbb{E}\left[\phi ({z}^{t+1})-\phi^*\right] +\alpha\mathcal{E}^{t+1} \\
\le& \left(1-\frac{\alpha\mu}{4}\right) \left[\mathbb{E}\left[\phi ({z}^{t})-\phi^*\right]+\alpha\mathcal{E}^{t}\right] \\
&+ \frac{\alpha(6+\delta)(1+3\beta^2K^3L^2)\sigma^2}{BK}+\frac{\alpha B_h}{8}.
\end{align*}
By recursively substituting this inequality, we can derive the desired result in \eqref{equ:pl_convergence}:
\begin{align*}
&\mathbb{E}\left[\phi ({z}^{T})-\phi^*\right] +\alpha\mathcal{E}^{T} \\
\le& \left(1-\frac{\alpha\mu}{4}\right)^{T} \left[\phi ({z}^{0})-\phi^* +\alpha\mathcal{E}^{0}\right]\\
&+ \left[\sum_{t=0}^{T-1} \left(1-\frac{\alpha\mu}{4}\right)^{t}\right]\left[\frac{\alpha\sigma^2(6+\delta)(1+3\beta^2K^3L^2)}{BK}+\frac{\alpha B_h}{8}\right] \notag \\
\le&\left(1-\frac{\alpha\mu}{4}\right)^{T} \left[\phi ({z}^{0})-\phi^* +\alpha\mathcal{E}^{0}\right] \\
&+ \frac{4(6+\delta)(1+3\beta^2K^3L^2)\sigma^2}{\mu BK}+\frac{B_h}{2\mu} \\
\overset{\eqref{equ:pl_condition}}{\le}&\left(1-\frac{\alpha\mu}{4}\right)^{T} \left[\phi ({z}^{0})-\phi^* +\alpha\mathcal{E}^{0}\right] + \frac{25\sigma^2}{\mu BK}+ \frac{\sigma^2}{16\mu B}+\frac{B_h}{2\mu},
\end{align*}
where the second inequality is obtained by using the fact that $\sum_{t=0}^{\infty} \gamma^{t}=\frac{1}{1-\gamma}$, for $\gamma\in(0,1)$.

\end{appendices}

\bibliographystyle{ieeetr}
\bibliography{ref}

\end{document}